\documentclass[10pt,twocolumn,letterpaper]{article}

\usepackage[pagenumbers]{cvpr} 
\usepackage{xcolor,colortbl}
\usepackage{color}
\usepackage{comment}
\usepackage{booktabs}
\usepackage{pifont}
\usepackage{amsfonts}
\usepackage{bbm}
\usepackage{soul}
\usepackage{float}
\usepackage{multirow}
\usepackage{array,makecell}
\usepackage{arydshln}
\usepackage{textcomp}
\usepackage{subcaption}
\usepackage{ifthen}
\usepackage{graphicx}

\newcommand{\cmark}{\ding{51}}%
\newcommand{\xmark}{\ding{55}}%

\definecolor{cvprblue}{rgb}{0.21,0.49,0.74}
\usepackage[pagebackref,breaklinks,colorlinks,citecolor=cvprblue]{hyperref}

\title{Gen2Det: Generate to Detect}

\author{
Saksham Suri$\text{}^{1}$\thanks{Work done during internship at Meta.}
\and
Fanyi Xiao$\text{}^{2}$
\and
Animesh Sinha$\text{}^{2}$
\and
Sean Chang Culatana$\text{}^{2}$\vspace{0.01in}
\and 
Raghuraman Krishnamoorthi$\text{}^{2}$
\and
Chenchen Zhu$\text{}^{2}$\thanks{*Equal Advisory Contribution.}
\and
Abhinav Shrivastava$\text{}^{1}$\footnotemark[2]\vspace{0.01in}
\and
{University of Maryland, College Park$^{1}$}\quad\quad\quad
{Meta$^{2}$}
}

\begin{document}
\maketitle

\begin{abstract}
Recently diffusion models have shown improvement in synthetic image quality as well as better control in generation. We motivate and present Gen2Det, a simple modular pipeline to create synthetic training data for object detection for free by leveraging state-of-the-art grounded image generation methods. Unlike existing works which generate individual object instances, require identifying foreground followed by pasting on other images, we simplify to directly generating scene-centric images. In addition to the synthetic data, Gen2Det also proposes a suite of techniques to best utilize the generated data, including image-level filtering, instance-level filtering, and better training recipe to account for imperfections in the generation. Using Gen2Det, we show healthy improvements on object detection and segmentation tasks under various settings and agnostic to detection methods. In the long-tailed detection setting on LVIS, Gen2Det improves the performance on rare categories by a large margin while also significantly improving the performance on other categories, e.g. we see an improvement of 2.13 Box AP and 1.84 Mask AP over just training on real data on LVIS with Mask R-CNN. In the low-data regime setting on COCO, Gen2Det consistently improves both Box and Mask AP by 2.27 and 1.85 points. In the most general detection setting, Gen2Det still demonstrates robust performance gains, e.g. it improves the Box and Mask AP on COCO by 0.45 and 0.32 points.
\end{abstract}
    
\section{Introduction}
\label{sec:intro}

\begin{figure}[!t]
\centering
\includegraphics[width=\linewidth]{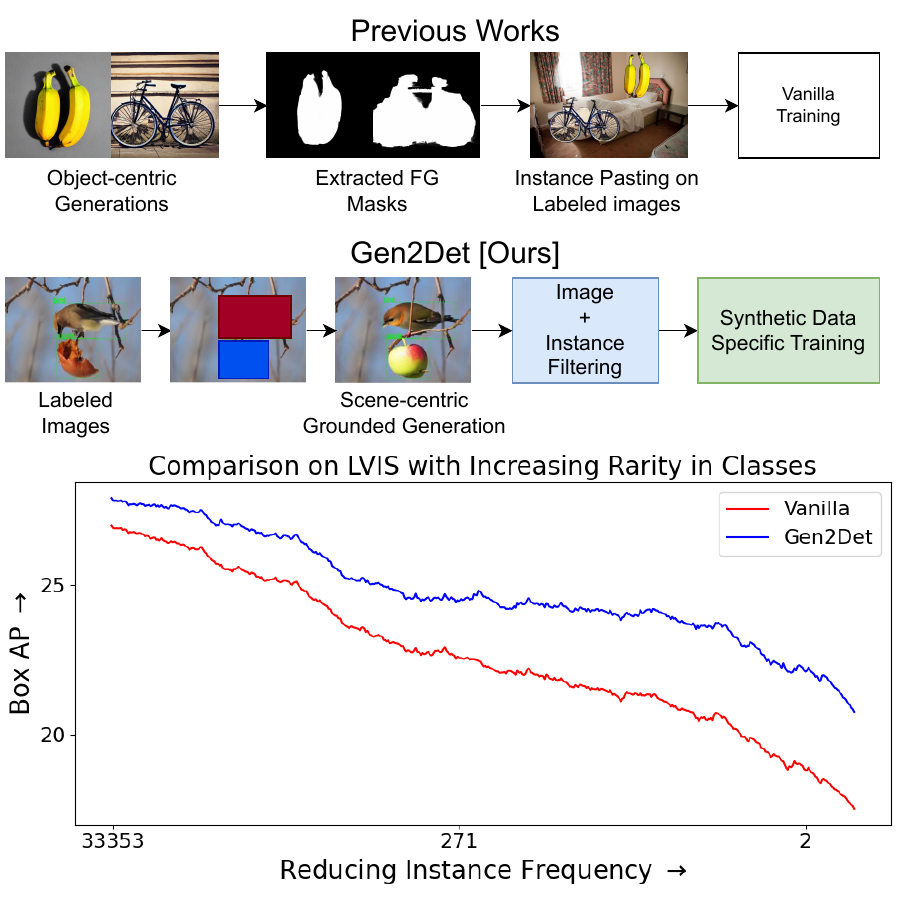}
  \caption{
  Existing approaches which utilize synthetic data for detection training follow a common methodology of generating object-centric images and pasting instances on real images (top). Gen2Det (middle) instead utilizes state-of-art grounded inpainting diffusion model to directly generate scene-centric images. Further, Gen2Det performs filtering to handle unsuitable generations. Finally, during the detector training we introduce changes to handle filtered generated data better. As a result, our method consistently improves over vanilla training and the AP improvements increase (bottom) as the class becomes rare (\ie, long-tailed classes).
  }
\label{fig:teaser}
\vspace{-0.2in}
\end{figure}

Recent developments in generative modelling using diffusion models has drastically improved the generation quality. 
Works like LDM~\citep{rombach2022high}, DALL-E~\citep{ramesh2021zero, ramesh2022hierarchical}, Imagen~\citep{saharia2022photorealistic}, Parti~\citep{yu2022scaling} 
have shown the generation power which diffusion models possess. In addition to these models which can generate high quality images given a text input, there have been multiple developments in the direction of higher control in generation. Along this direction exist works which use conditional control like ControlNet~\citep{zhang2023adding} and GLIGEN~\cite{li2023gligen}. There are also works like LoRA~\citep{hu2021lora} and Dreambooth~\cite{ruiz2023dreambooth} which provide means to adapt these large models in a quick and efficient manner to generate specific kinds of images. With these developments in both quality and control of synthetically generated images, it is only natural to come back to the question of ``How to best utilize data generated from these models for improving recognition performance?''.

Most previous works~\citep{he2022synthetic, sariyildiz2023fake, tian2023stablerep} have explored using synthetic data from diffusion models for classification and pre-training tasks. The goal of our work is to look at utilizing more realistic scene configurations by generating images conditioned on the existing layout of boxes and labels and use them for object detection and segmentation. 
A recent work, XPaste~\citep{zhao2023x}, explores this problem and utilizes techniques similar to simple copy-paste~\citep{ghiasi2021simple} to paste synthetically generated object-centric instances onto real images for object detector training. 
However, their method relies on off-the-shelf segmentation methods built over CLIP~\citep{radford2021learning, yun2022selfreformer, Qin_2020_PR, 2203.04708, lueddecke22_cvpr} to extract masks, and is thus subject to segmentation errors. 
In addition to the extra components and compute, the generated images from XPaste are also not realistic as they do not respect natural layouts due to random pasting as shown in Figure~\ref{fig:teaser} (top).
In contrast, as shown in Figure~\ref{fig:teaser} (middle), Gen2Det leverages state-of-art diffusion models for grounded inpainting to generate scene-centric images which look more realistic.
The goal of our approach is to show that utilizing such generated data from state-of-art diffusion models can lead to improvement in performance of object detection and segmentation models. 
By using a grounded inpainting diffusion model as our base generator, we are able to generate synthetic versions of common detection datasets like LVIS~\citep{gupta2019lvis} and COCO~\citep{lin2014microsoft} in a layout conditioned manner. 
We then carefully design a set of filtering and training strategies, with which we demonstrate that we can improve detection performance when training on the joint set of real and synthetic images. More specifically, we sample batches of synthetic and real images with a sampling probability. During loss computation we also modify the loss for synthetic data to account for filtered instances. Additionally, without utilizing any additional models or data (including segmentation masks) during training, we show improvements in segmentation performance as a byproduct. The clear improvement over vanilla training shown in Figure~\ref{fig:teaser} (bottom) for classwise Box AP with increasing rarity of classes makes a strong case for such a pipeline especially in long-tailed or low data regimes.

We emphasize the importance of the proposed filtering and training techniques to incorporate synthetic data through our ablations, as we show that directly training on the generated data or mixing it with real data in a naive manner ends up hurting the performance. It is important to do the proper filtering and loss modifications to mitigate shortcomings of the generations and let the model learn from the mix of real and synthetic data.
We propose Gen2Det as a general approach to utilize synthetic data for detector training. 
Due to its modular nature, different components can be updated as the field progresses. This includes utilizing better generators, filtering techniques, model architectures and also training recipes. 
We highlight our contributions below:

\begin{itemize}
    \item Propose to use state-of-art grounded-inpainting models to generate synthetic data in a manner that respects the realistic scene layouts. 
    \item Propose techniques to perform filtering at image and instance level along with changes in the detector training in order to utilize such synthetic data effectively and handle imperfections in generated instances. 
    \item Through our experiments we show consistent improvement across datasets and architectures especially in long-tailed and low data settings. There is also considerable gains for rare categories.
    \item We also provide quantitative and qualitative ablations to showcase the effect of different hyperparameters and components of our approach which would help further research in usage of such synthetically generated data.
\end{itemize}
\begin{figure*}[!t]
\centering
\includegraphics[width=\linewidth]{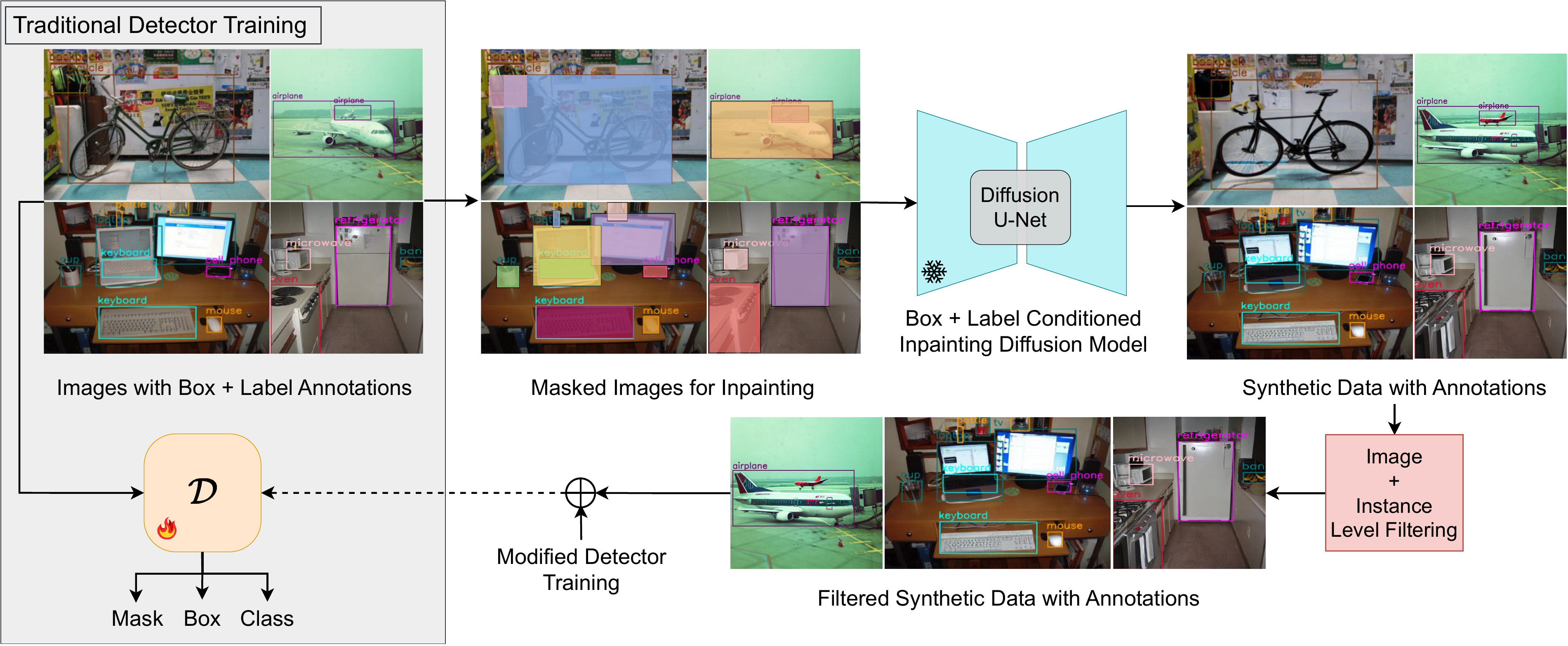}
  \caption{
  \textbf{Gen2Det: our proposed pipeline for generating and utilizing synthetic data for object detection and segmentation.} Gen2Det starts by generating grounded inpainted images using state-of-art diffusion model. The generated images are then filtered at image and instance level to remove globally bad images as well as individual instances which might be low quality or have hallucinations. 
  Finally, we train object detection and segmentation models using the filtered data along with our improved training methodology by introducing sampling and background ignore.
 }
\label{fig:approach}
\vspace{-0.2in}
\end{figure*}
\section{Related Works}
\label{sec:related_works}

\noindent\textbf{Diffusion Models for Controllable Image Generation.} 
Diffusion models have been shown to generate images with unprecedented high quality. 
Amongst these models a few of the popular models which can generate images from text prompts include LDM~\cite{rombach2022high}, DALL-E~\cite{ramesh2021zero, ramesh2022hierarchical}, Imagen~\cite{saharia2022photorealistic}, and Parti~\cite{yu2022scaling}. 
In addition to just text guided generation there have been further improvements in having more controllable generation and editing through different inputs~\cite{saharia2022palette, voynov2023sketch, brooks2023instructpix2pix, meng2021sdedit,hollein2023text2room,patashnik2023localizing,zhang2023adding,li2023gligen}.
One of the popular works ControlNet~\cite{zhang2023adding} provides multiple ways to enable control including using edge maps, scribbles, segmentation masks amongst other modalities. 
Another work GLIGEN~\cite{li2023gligen} brings in more grounding control where they show image generation and inpainting conditioned on boxes, keypoints, HED maps, edge maps and semantic maps. We also utilize a similar pre-trained state-of-art diffusion model for grounded inpainting. 

\noindent\textbf{Synthetic Data for Detection.} There has been prior non-diffusion based work exploring the use of synthetic data for detection. 
Some works have explored using image-based blending all the way to depth and semantics informed positioning~\cite{georgakis2017synthesizing} for creating synthetic data. 
There have been other works which use computer graphics based approaches to render synthetic data by varying lighting and orientations~\cite{Tremblay2018TrainingDN,Rajpura2017ObjectDU}. 
Further there have been a line of augmentations belonging to the copy-paste family~\cite{ghiasi2021simple,Dwibedi2017CutPA,Fang2019InstaBoostBI,Dvornik2018ModelingVC} where object instances are cropped and pasted on images at different locations. All these approaches end up repeating instances already present in the training set.

\noindent\textbf{Leveraging Synthetic Data from Diffusion Models.} 
Due to its high quality generation and the flexibility in handling multimodal data (\eg, vision and language), there has been a lot of recent work in using pretrained diffusion models for different tasks.
Peekaboo~\cite{burgert2022peekaboo} shows the use of diffusion models as zero shot segmentors while other works~\cite{li2023your,clark2023text} shows that diffusion models can act as zero shot classifiers. 
StableRep~\cite{tian2023stablerep} on the other hand uses data generated by diffusion models to train self-supervised models. 
A few method also explore using diffusion model features~\cite{li2023dreamteacher,mukhopadhyay2023diffusion,tang2023emergent,luo2023dhf,xiang2023denoising} directly for downstream tasks. There have also been works exploring the use of synthetic data from diffusion models to improve image classification~\cite{he2022synthetic, sariyildiz2023fake, azizi2023synthetic, bansal2023leaving,hemmat2023feedback}. 
While the classification task has had a lot of exploration with synthetic data due to the object-centric nature of images which these models can generate easily, 
the detection task is less explored as it's harder to generate data with grounded annotations using these models. 
A recent study~\cite{lin2023explore} explored the use of such data in highly constrained few shot settings for detection where the gains are expected. 
Recently, XPaste~\cite{zhao2023x} looked at more general benchmarks and showed consistent improvements using the Centernet2~\cite{zhou2021probabilistic} architecture. 
Specifically, XPaste~\cite{zhao2023x} uses diffusion models to generate object-centric images and uses multiple CLIP~\cite{radford2021learning} based approaches~\cite{yun2022selfreformer, Qin_2020_PR, 2203.04708, lueddecke22_cvpr} to extract segmentation maps which makes it slow and error-prone. 
Following the extraction of instances and their segmentation maps, they use synthetic and real instances (retrieved from an additional real dataset) to perform copy-paste~\cite{ghiasi2021simple} augmentation to generate synthetic data. Gen2Det on the other hand does not use any additional CLIP based approaches. Rather, once we generate the data we only train with bounding box labels but show improvements in both box and mask AP.  Also in terms of training speed, we are $3.4\times$ faster compared to XPaste with the same configuration.

\section{Approach}
\label{sec:aproach}

\subsection{Overview}
Through Gen2Det, we provide a modular way to generate and utilize synthetic data for object detection and instance segmentation. 
As shown in Figure~\ref{fig:approach}, we start by generating data using state-of-art grounded image inpainting diffusion model. 
As the generations may not always be perfect we perform image level and instance level filtering. 
The image level filtering is performed using a pre-trained aesthetic classifier~\cite{Christophschuhmann} while the instance level filtering is performed using a detector trained on the corresponding real data. 
After performing filtering, the images are suitable for training using our proposed training strategy. 
While we do not make any architectural changes to the detector to keep the pipeline general enough, we perform a probability based sampling of a batch to be composed of synthetic or real samples. 
Further while computing the losses, we modify the negatives corresponding to the synthetic data to be ignored from loss computation to account for the filtering we performed. We also do not apply mask loss for synthetic samples as we do not have the segmentation masks corresponding to them.
Following these steps we are able to show consistent improvements in performance.

\begin{table*}
\begin{tabular}{ll}
     \begin{minipage}[t]{0.64\linewidth}
 \setlength{\cmidrulewidth}{0.01em}
\renewcommand{\tabcolsep}{8pt}
\renewcommand{\arraystretch}{1.1}
\resizebox{\linewidth}{!}{
\begin{tabular}{@{}lcccccccc@{}}
\toprule
 \multirow{2}{*}{Method} &  \multicolumn{4}{c}{Box}&  \multicolumn{4}{c}{Mask}\\
 \cmidrule[\cmidrulewidth](l){2-5}
 \cmidrule[\cmidrulewidth](l){6-9}
 &  $\text{AP}$ & $\text{AP}_r$ & $\text{AP}_c$ & $\text{AP}_f$ &  $\text{AP}$ & $\text{AP}_r$ & $\text{AP}_c$ & $\text{AP}_f$ \\
 \midrule 
 Vanilla (real only) & 33.80	&20.84	&32.84	&40.58	&29.98	&18.36	&29.64	&35.46\\
 Vanilla (synth only) & 11.27 & 6.54 & 9.35 & 15.51 & - & - & - & -\\
 Vanilla (synth+real) & 31.82 & 20.68 & 31.18 & 37.42 & 27.49 & 18.35 & 27.08 & 31.96\\
 XPaste &34.34	&21.05	&33.86	&40.71&	30.18&	18.77&	30.11 &	35.28\\
 Ours &\textbf{34.70}	&\textbf{23.78}&	\textbf{33.71}&	\textbf{40.61}&	\textbf{30.82}	&\textbf{21.24}	&\textbf{30.32}	&\textbf{35.59}\\
\bottomrule
\end{tabular}}
\caption{Comparisons on LVIS with baselines using Centernet2 architecture. We report the overall AP and also AP for rare ($\text{AP}_r$), common ($\text{AP}_c$) and frequent ($\text{AP}_f$) classes for both box and mask evaluations.}\label{tab:lvis_main}
\end{minipage}
     & 
\begin{minipage}[t]{0.3\linewidth}
\setlength{\cmidrulewidth}{0.01em}
\renewcommand{\tabcolsep}{10pt}
\renewcommand{\arraystretch}{1.1}

\resizebox{\linewidth}{!}{
\begin{tabular}{@{}lcc@{}}
\toprule
 Method &  $\text{AP}_b$ & $\text{AP}_m$ \\
 \midrule
  Vanilla (real only) & 	46.00	& 39.8\\
 Vanilla (synth only) & 24.51 & - \\
 Vanilla (synth+real) & 44.35 & 37.03 \\
 XPaste &	46.60	& 39.90\\
 Ours &\textbf{47.18}	& \textbf{40.44}\\
\bottomrule
\end{tabular}
}
\caption{Comparisons with baselines on COCO dataset using Centernet2 architecture. We report the Box AP ($\text{AP}_b$) and Mask AP ($\text{AP}_m$).}\label{tab:coco_main}
\end{minipage}
\end{tabular}
\vspace{-0.1in}
\end{table*}

\subsection{Image Generation}
We start our pipeline by generating synthetic images by utilizing a state-of-art grounded inpainting diffusion model.
This model is trained to support multiple kinds of input conditions. 
We utilize the model trained for image inpainting with the image, boxes, and corresponding labels as input. 
Specifically, for each image $\mathcal{I}$ in the dataset with box annotations $\mathcal{B}$, we provide as input the image and the corresponding annotations to the diffusion model and ask it to inpaint the annotated regions with new instances of the same class by providing the box label as input. 
As inpainting model requires an image and box level text description, for each box $b_i$ we use the class name $\textless c_i\textgreater$ as the box level prompt and a concatenation of strings $\textless \text{a } c_1, \text{a } c_2, \text{ ... and a } c_n\textgreater$ as the image level prompt where $n$ is the number of instances in the image. 
We show examples of images generated using this technique in Figure~\ref{fig:generations}.

\subsection{Filtering}
The images generated using the strategy described above may not always contain good generations so it is important to filter out low quality generations before using them to train a detection model. We perform two levels of filtering. First we filter at image level followed by a more granular instance level filtering. 
We describe each of these below.

\subsubsection{Image Level Filtering}
The first level of filtering we perform is an image level filtering using a pretrained classifier which utilizes a CLIP+MLP architecture as described in~\citep{Christophschuhmann}. 
The model is trained to predict how pleasing the image looks visually by assigning an aesthetic score to each image. 
We utilize the publicly available weights from~\citep{Christophschuhmann} and pass every generated image through the classifier which returns an aesthetic score for each image. 
Based on qualitatively analyzing images and computing the average aesthetic score for real images (from COCO) we set a threshold for aesthetic filtering $\tau_a$. 
Any image with an aesthetic score less than $\tau_a$ is discarded and its annotations are removed. 
We show the effect of this filtering by visualizing discarded samples in Figure~\ref{fig:aesthetic}. 

\subsubsection{Detector Filtering}
As a more granular level of quality assurance, we also perform an extra filtering step at the instance level. 
This filtering is designed to remove annotations for specific generated instances which do not have good generation quality. 
In order to perform this filtering we first train a detector on only the real data. 
We then pass all the generated images through the trained detector and store its predictions. 
Based on the detectors predictions we evaluate whether a ground truth annotation corresponding to an inpainted region should be utilized for training or not. 
This step is important to handle poor quality/incorrect generations. 
To do this, for each generated image we iterate over all ground truth annotations used to generate it. 
For each annotation we go through all predictions and remove the ground truth annotation if there is no overlapping prediction with a score greater than $\tau_s$ and IoU greater than $\tau_{iou}$. 
This helps us remove ground truth annotations corresponding to generation which are either bad quality as the pre-trained detector is not able to predict it with even a low confidence threshold, or for which the region where they are generated is not the ground truth region. 
We show results of the kind of annotations removed using this filtering in Figure~\ref{fig:det}.

\begin{figure*}
\centering
\includegraphics[width=\linewidth]{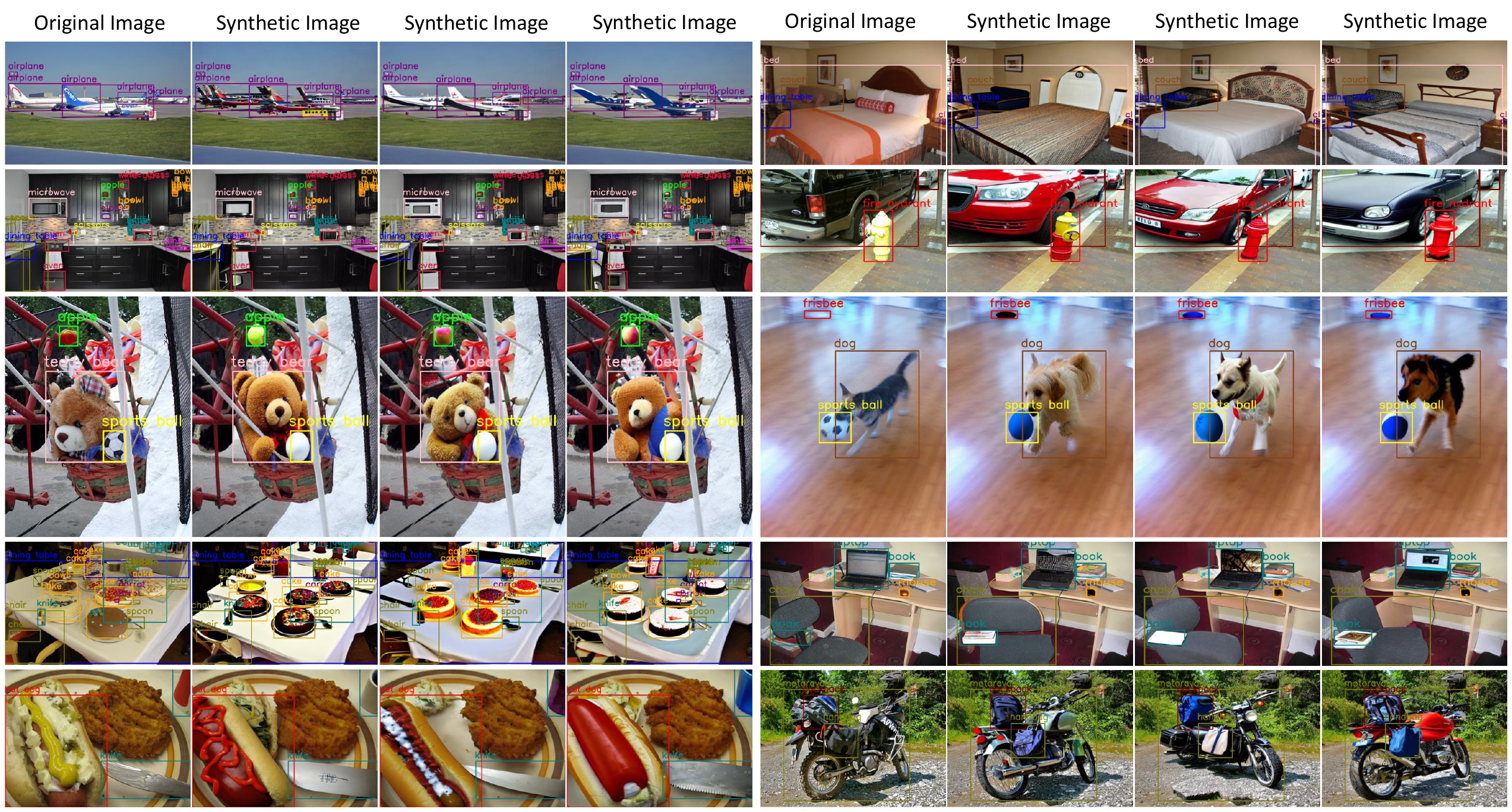}
  \caption{
  Examples of generations using the inpainting diffusion model on the COCO dataset. 
  The first and fifth columns correspond to the original COCO images and the rest of the columns are generations with different seeds.
  These generated images are then fed to our filtering pipeline for further processing before use in training. 
  }
\label{fig:generations}
\vspace{-0.2in}
\end{figure*}

\subsection{Model Training}
Once the data is ready for use we describe our training strategy to utilize the synthetic dataset along with the real dataset. 
It should be noted that for synthetically generated images we do not have the segmentation masks for the instances so we do not apply the mask loss on those images.

\subsubsection{Batch Sampling}
We utilize both the real and synthetic data for training the final model. 
To do this, we need to figure out how to mix the two datasets together in a way which lets us utilize the full potential of the real data but at the same time extract as much useful information from the additional synthetic data as possible. 
Naively using only the synthetic dataset and a naive combination of synthetic and real datasets during training leads to drop in performance as shown in the ablations.
Therefore, to best use both the real and synthetic datasets, we end up defining a sampling probability $p$. 
Specifically, a batch is chosen to comprise of completely synthetically generated samples with the sampling probability $p$ during training. 
This sampling strategy interleaves batches of synthetic and real data and as we will show leads to effective learning from both data sources. 

\subsubsection{Background Ignore}
While the filtering we perform at both image and instance level can deal with bad quality generations, they also introduce some noise especially for classes for which the detector itself has poor quality predictions. 
Essentially, the detector filtering removes ground truth annotation corresponding to bad quality/incorrect generations but that does not remove the bad instance itself from the image.
Additionally, the generative model could have also hallucinated multiple instance of an object class leading to missing annotations. 
To counter the effect of both these scenarios we introduce an ignore functionality during training. 
Essentially, for both the region proposal network (RPN) and detector head we ignore the background regions from the loss computation, if their fg/non-bg class prediction score is higher than a threshold $\tau_i$. 
This lets us train even in the presence of some bad quality regions without incorrectly penalizing the detector for predicting objects at those locations. 
This is an important change in training to allow our model to effectively utilize synthetic images with incorrectly hallucinated instances, as the instance level filtering only removes the instance annotation but does not handle hallucinations. 

Additionally, as stated above we do not utilize any additional mask information for the synthetic data as we only inpaint using boxes.  So while training we ignore the mask loss for the synthetic images. 
Even with this, as shown in our quantitative results, we see an improvement in both box and mask performance.


\section{Results}
\label{sec:experiments}

\subsection{Experimental Setting}
We evaluate our approach on LVIS~\cite{gupta2019lvis} and COCO~\cite{lin2014microsoft} datasets. 
These datasets provide a standard benchmark for object detection and instance segmentation approaches. 
LVIS is a long tailed dataset with 1203 classes utilizing the same images as COCO which contains only 80 classes. 
The long tailed nature of LVIS makes it especially appealing to showcase the use of synthetic data for the rare categories. 
We report the Average Precision (AP) for both these datasets and compare using the box AP ($\text{AP}_b$) as well as mask AP ($\text{AP}_m$). 
Additionally for LVIS, following standard practice and the existing definition of rare, common, and frequent classes, we report the respective box and mask APs for those subset of classes. 
Further, we also report results on artificially created subsets of COCO to simulate the low-data regime by using $1\%$, $5\%$, $10\%$, $20\%$, $30\%$, $40\%$ and $50\%$ randomly selected images from COCO.

\begin{figure}
\centering
\includegraphics[width=\linewidth]{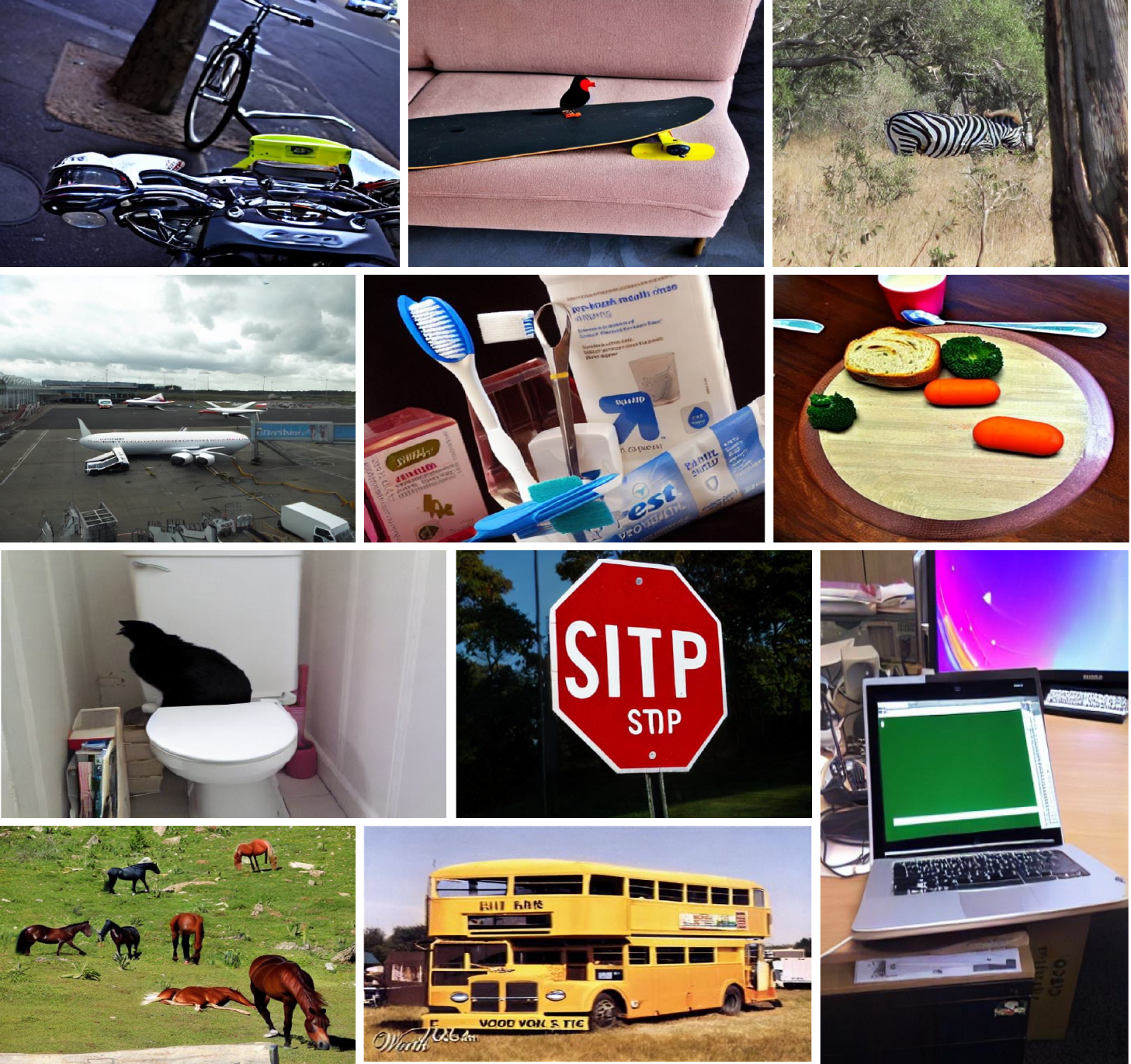}
  \caption{
  Examples of samples discarded during the image level filtering. 
  As can be seen, image level filtering is able to remove images with artifacts present at a global level.
  }
\label{fig:aesthetic}
\vspace{-0.2in}
\end{figure}

We utilize the Detectron2 framework~\cite{wu2019detectron2} for training all the models and use the XPaste code to run their baselines.
We use the Centernet2 clone adapted for instance segmentation. 
For LVIS, we rerun XPaste to get the results corresponding to adding only additional synthetic data. 
For COCO, we directly adopt their evaluation numbers due to the lack of corresponding configs and details for reproduction. 
Following their code we also utilize the ImageNet-22k~\cite{Deng2009ImageNetAL} pretrained models for the Centernet2 experiments while using the ImageNet-1k~\cite{Deng2009ImageNetAL} standard models from Detectron2 for the rest of the experiments. 
We utilize the Mask R-CNN~\cite{He2017MaskR} architecture and the LVIS dataset for our ablations. 
The Mask R-CNN~\cite{He2017MaskR} and Faster R-CNN~\cite{Ren2015FasterRT} are implemented in the Detectron2 framework~\cite{wu2019detectron2}.

\begin{figure*}
\centering
\includegraphics[width=\linewidth]{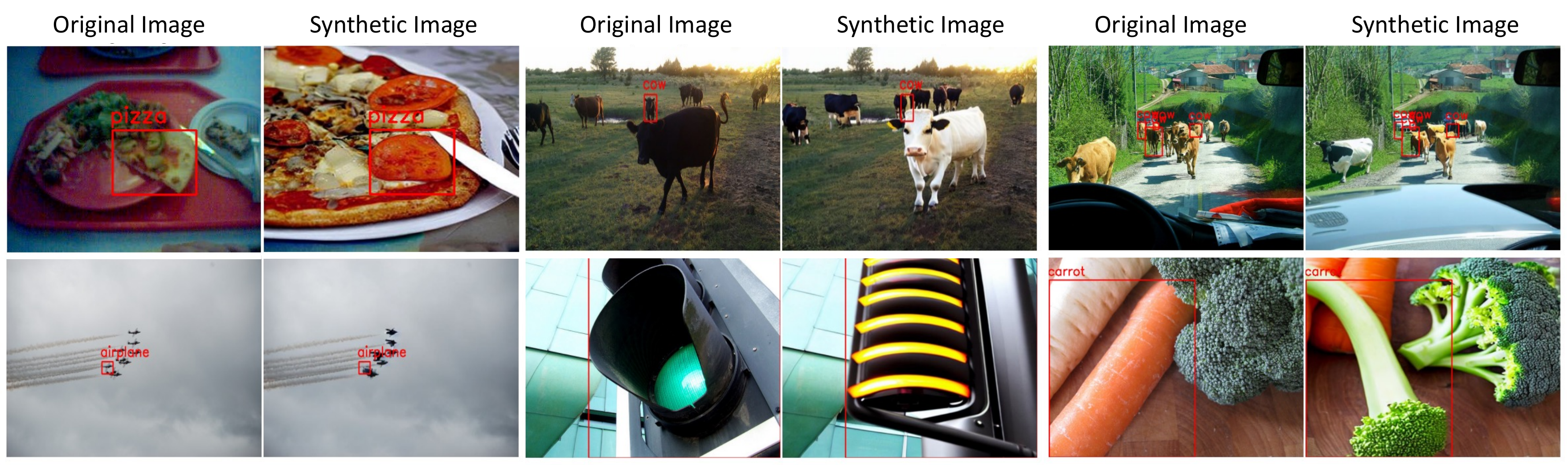}
  \caption{Examples of ground-truth instance annotations discarded by the detector filtering highlighted in red.}
\label{fig:det}
\vspace{-0.1in}
\end{figure*}

For data generation, we utilize a state-of-art diffusion model for grounded inpainting using the images from LVIS and COCO along with their annotations which contain the boxes and class labels. 
For aesthetic filtering, we utilize the open source model available in their repository~\cite{Christophschuhmann}. 
For aesthetic filtering we set $\tau_a$ to $4.5$ which is roughly the same as the average aesthetic score on real COCO images. 
For detector filtering we set $\tau_s$ as $0.2$ and $\tau_{iou}$ as $0.3$ for LVIS. 
We use the same $\tau_{iou}$ for COCO and set $\tau_s$ as $0.1$. 
During training, we use a sampling probability $p = 0.2$. We set $\tau_i$ to 0 for both datasets thus effectively ignore all background regions corresponding to the synthetic data from the loss.

\begin{table*}[t]
 \setlength{\cmidrulewidth}{0.01em}
\renewcommand{\tabcolsep}{12pt}
\renewcommand{\arraystretch}{1.0}
\centering
\resizebox{0.85\linewidth}{!}{
\begin{tabular}{@{}lcccccccc@{}}
\toprule
 \multirow{2}{*}{Method} &  \multicolumn{4}{c}{Box}&  \multicolumn{4}{c}{Mask}\\
 \cmidrule[\cmidrulewidth](l){2-5}
 \cmidrule[\cmidrulewidth](l){6-9}

 &  $\text{AP}$ & $\text{AP}_r$ & $\text{AP}_c$ & $\text{AP}_f$ &  $\text{AP}$ & $\text{AP}_r$ & $\text{AP}_c$ & $\text{AP}_f$ \\
 \midrule
 Faster R-CNN &21.39	&10.05	&19.46	&28.51 &-&-&-&-\\
 Faster R-CNN (Ours)& \textbf{22.90}	&\textbf{12.78}	&\textbf{21.19}	&\textbf{29.27}&-&-&-&-\\
 \hdashline
 Mask R-CNN  & 22.29 & 10.63 & 20.15 & 29.80 & 21.83 & 11.15 & 20.42 & 28.10 \\
 Mask R-CNN (Ours) &  \textbf{24.42} & \textbf{15.43} &\textbf{ 22.63} & \textbf{30.38} & \textbf{23.67} &\textbf{ 15.33} & \textbf{22.62} & \textbf{28.51}\\
 \hdashline
 Centernet2 &33.80	&20.84	&32.84	&40.58	&29.98	&18.36	&29.64	&35.46\\
 Centernet2 (Ours) &\textbf{34.70}	&\textbf{23.78}&	\textbf{33.71}&	\textbf{40.61}&	\textbf{30.82}	&\textbf{21.24}	&\textbf{30.32}	&\textbf{35.59} \\
 
\bottomrule
\end{tabular}
}
\caption{
Results across different backbones on the LVIS dataset. We can see consistent improvement across various backbones with especially higher gains on the rare categories.}
\label{tab:architectures} %
\vspace{-0.15in}
\end{table*}

\begin{table}
 \setlength{\cmidrulewidth}{0.01em}
\renewcommand{\tabcolsep}{10pt}
\renewcommand{\arraystretch}{1.0}
\centering
\resizebox{\linewidth}{!}{
\begin{tabular}{@{}ccccc@{}}
\toprule
 \multirow{2}{*}{Image \%} &  \multicolumn{2}{c}{Vanilla}&  \multicolumn{2}{c}{Gen2Det}\\
 \cmidrule[\cmidrulewidth](l){2-3}
 \cmidrule[\cmidrulewidth](l){4-5}

 &  $\text{AP}_b$ & $\text{AP}_m$ & $\text{AP}_b$ & $\text{AP}_m$ \\
 \midrule
 1	& 11.76	&11.34	&14.84 (\textcolor{blue}{+3.08})	&13.89 (\textcolor{blue}{+2.55}) \\
2	&16.07	&15.36	&18.42 (\textcolor{blue}{+2.35})	&17.27 (\textcolor{blue}{+1.91}) \\
5	&19.52	&18.48	&22.91 (\textcolor{blue}{+3.38})	&21.17 (\textcolor{blue}{+2.69}) \\
10	&25.29	&23.35	&27.62 (\textcolor{blue}{+2.33})	&25.34 (\textcolor{blue}{+1.99}) \\
20	&29.32	&26.97	&31.76 (\textcolor{blue}{+2.44})	&29.03 (\textcolor{blue}{+2.06}) \\
30	&33.02	&30.37	&34.70 (\textcolor{blue}{+1.68})	&31.65 (\textcolor{blue}{+1.28}) \\
40	&34.39	&31.36	&36.09 (\textcolor{blue}{+1.69})	&32.67 (\textcolor{blue}{+1.30}) \\
50	&36.52	&33.20	&37.74 (\textcolor{blue}{+1.22})	&34.24 (\textcolor{blue}{+1.04}) \\
\bottomrule
\end{tabular}
}
\caption{
Performance on the LD-COCO setting. The LD-COCO data is a randomly selected subset of COCO with different percentages of images from the dataset.
}\label{tab:ld_coco} %
\vspace{-0.2in}
\end{table}

\begin{table*}[!ht]
 \setlength{\cmidrulewidth}{0.01em}
\renewcommand{\tabcolsep}{3pt}
\renewcommand{\arraystretch}{1.1}
\centering
\begin{tabular}{@{}ccccccccccc@{}}
\toprule
Row& Real Data & Synth. Data & Sampling & Img. Filt. & Det. Filt. & Bg. Ignore & Box AP & $\Delta_{\text{AP}}$ & Box $\text{AP}_{r}$ & $\Delta_{\text{AP}_r}$   \\
 \midrule
\rowcolor{gray!0}1&\cmark & \xmark & \xmark & \xmark & \xmark & \xmark & 22.29& $-$ & 10.63 & $-$\\
\rowcolor{gray!5}2&\xmark & \cmark & \xmark & \xmark & \xmark & \xmark & 7.67 & -15.25 & 4.57 & -6.06\\
\rowcolor{gray!5}3&\cmark &\cmark & \xmark & \xmark & \xmark & \xmark & 20.75& -1.54 & 10.91 & 0.28\\
\rowcolor{gray!5} 4&\cmark &\cmark & \cmark & \xmark & \xmark & \xmark & 23.49& 1.2 & 13.21 & 2.58\\
\rowcolor{gray!15}5&\cmark &\cmark & \cmark & \cmark & \xmark & \xmark & 23.78& 1.49 & 13.79 & 3.16\\
\rowcolor{gray!15}6&\cmark &\cmark & \cmark & \cmark & \cmark & \xmark & 23.76& 1.47 & 13.62 & 2.99\\
\rowcolor{gray!15}7&\cmark &\cmark & \cmark & \cmark & \xmark & \cmark & 23.32& 1.03 & 12.77 & 2.14\\
\rowcolor{gray!30}8&\cmark &\cmark & \cmark & \cmark & \cmark & \cmark & 24.42& 2.13 & 15.43& 4.8\\
\bottomrule
\end{tabular}
\caption{Ablation of various components of proposed approach on LVIS using Mask R-CNN.}
\label{tab:ablation_all}
\vspace{-0.1in}
\end{table*}

\begin{table*}[!h]
  \begin{subfigure}[b]{0.245\linewidth}
    \renewcommand{\tabcolsep}{4pt}
    \renewcommand{\arraystretch}{1.5}
    \centering
    \resizebox{\linewidth}{!}{
      \begin{tabular}{@{}lcccc@{}}
        \toprule
        Config & $1\times$ & $2\times$ & $3\times$ & $4\times$ \\
        \midrule
        AP & 22.50 & 24.42 & 22.85 & 22.31 \\
        \bottomrule
      \end{tabular}
    }
    \caption{Longer training schedule.}
    \label{tab:long}
  \end{subfigure}%
  \hfill
  \begin{subfigure}[b]{0.355\linewidth}
    \renewcommand{\tabcolsep}{3pt}
    \renewcommand{\arraystretch}{1.5}
    \centering
    \resizebox{\linewidth}{!}{
      \begin{tabular}{@{}lcccccc@{}}
        \toprule
        Probability & $0.0$ & $0.1$ & $0.2$ & $0.3$ & $0.4$ & $0.5$ \\
        \midrule
        AP & 22.29 & 23.23 & 24.42 & 23.42 & 23.28 & 22.52 \\
        \bottomrule
      \end{tabular}
    }
    \caption{Effect of sampling probability.}
    \label{tab:probability}
  \end{subfigure}%
  \hfill
  \begin{subfigure}[b]{0.37\linewidth}
    \renewcommand{\tabcolsep}{3pt}
    \renewcommand{\arraystretch}{1.55}
    \centering
    \resizebox{\linewidth}{!}{
      \begin{tabular}{@{}lcccccc@{}}
        \toprule
        Synthetic Data & $0\times$ & $1\times$ & $2\times$ & $3\times$ & $4\times$ & $5\times$ \\
        \midrule
        AP & 22.29 & 23.55 & 24.42 & 23.67 & 23.82 & 23.82 \\
        \bottomrule
      \end{tabular}
    }
    \caption{Effect of generated data sample counts.}
    \label{tab:data}
  \end{subfigure}
  \caption{Ablation on different hyperparameters on the LVIS dataset with Mask R-CNN backbone.}
  \vspace{-0.1in}
\end{table*}

\subsection{Quantitative Results}
\subsubsection{Comparison with Existing Works}
\noindent\textbf{Comparison on LVIS.} 
We start by comparing our approach to the existing works in Table~\ref{tab:lvis_main} on the LVIS dataset with CenterNet2 backbone. 
Over vanilla training we show that our method improves the Box and Mask AP over rare categories by $2.94$ and $2.88$ respectively with a $0.9$ and $0.84$ Box and Mask AP improvement overall. 
Despite the fact that XPaste utilizes four different CLIP based models~\cite{yun2022selfreformer, Qin_2020_PR, 2203.04708, lueddecke22_cvpr} to obtain segmentation masks and is also $~3.5\times$ slower to train compared to our approach on LVIS, 
we are able to outperform XPaste by $2.73$ Box AP and $2.47$ Mask AP on the rare categories and by $0.36 \text{ and } 0.64$ Box and Mask AP across all categories. 
The huge gains in rare categories and overall improvements highlight our methods effectiveness in both long tailed as well as general settings. 

\noindent\textbf{Comparison on COCO.} 
Similar to LVIS, we compare on the COCO benchmark in Table~\ref{tab:coco_main}. 
On COCO too we show an improvement of $1.18$ and $0.64$ on Box and Mask AP over vanilla training. 
Compared to XPaste we improve by $0.58$ Box AP and $0.54$ Mask AP. 
We note that compared to LVIS the improvements are slightly lower here as LVIS is a more long-tailed dataset where adding synthetic data shines.

\noindent\textbf{Comparison on LD-COCO.} 
In order to simulate a low-data regime on COCO with fewer images and labels, we create random subsets of different percentages of COCO images. 
In Table~\ref{tab:ld_coco} we show results on the low-data (LD) version of COCO. 
As can be seen for each row, we show consistent improvement in both Box and Mask AP. 
On average we improve the Box and Mask AP by $2.27$ and $1.85$ points respectively. This shows that adding synthetic data is especially fruitful in both long tailed and low data regimes.

\noindent\textbf{Comparison across backbones.} 
We further show comparison across architectures by comparing to performance with vanilla training using only the real data. 
As can be seen in Table~\ref{tab:architectures} we show consistent improvement in overall box and mask AP, and as mentioned above, our method especially shines for the rare categories. 
For the MaskRCNN architecture we show a substantial gain in performance with an improvement of $2.13$ Box AP and $1.84$ Mask AP over just training on real data.  
The improvements on rare categories are even higher with a gain of $4.8$ Box AP and $4.18$ Mask AP, highlighting the efficacy of using synthetic data in the Gen2Det pipeline for rare/long-tailed categories. 
For the same experimental setting on COCO, please see our Supplementary Materials.

\subsubsection{Ablation}

\noindent\textbf{Effect of different components.} 
Table~\ref{tab:ablation_all} summarizes the effect of incorporating different components in our pipeline. 
We report the Box AP and Box $\text{AP}_r$ for this analysis along with the respective improvement deltas introduced by each component over the baseline. 
Row 1 and 2 correspond to only training with the real and synthetic data respectively. 
It can be noted that naively training on only the synthetic or a simple combination of real and synthetic data (row 3) leads to a degradation in overall AP. 
Combining the two sets (row 3) does lead to a marginal increase in Box $\text{AP}_r$ due to the addition of more instances of the rare categories. 
Adding synthetic data along with sampling (row 4) gives us a boost of $1.2$ in overall Box AP and $2.58$ for the rare categories. 

Further, incorporating image level filtering which removes bad quality images in their entirety gives us a improvement of $1.49$ Box AP and $3.16$ Box $\text{AP}_r$ (row 5). 
Following our filtering pipeline, performing detector filtering on its own (row 6) does not work much better than just image level filtering as it ends up removing annotations corresponding to bad quality generations but the instances still remain in the images. 
Additionally we also try performing background loss ignore for the synthetic data without detector filtering (row 7), and that again does not lead to an improvement on its own rather leads to a drop in performance.
Finally, in row 8 we see that detector filtering coupled with background ignore takes care of the discarded annotations and leads to a net total improvement of $2.13$ AP and $4.8$ $\text{AP}_r$. 
It should be noted that detector filtering alone only removes certain annotations from training, 
using it with background ignore unlocks the full potential of such filtering and is important for training with synthetic data which might not always be perfect. 

\noindent\textbf{Effect of Longer training.} 
We try to see the effect of longer training by increasing the number of iteration for the detector training. 
We also adjust the learning rate steps proportionally for each experiment. 
As can be seen from Table~\ref{tab:long}, starting with the $1\times$ config and moving to $2\times$ gives us a boost as it is able to utilize the additional samples better but further longer training does not work that well and actually reduces performance as it starts overfitting to the data.

\noindent\textbf{Effect of Sampling Probability.} 
Here we vary the sampling probability $p$ used during training which affects whether the batch should be made from real or synthetic data. 
We show the change in performance for different synthetic probabilities in Table~\ref{tab:probability}. 
As can be seen in all cases we are better than the vanilla training. 
We do see that initially increasing the sampling probability to 0.2 helps and improves performance after which it decreases as the probability is further increased. 
This might be because the real data which is pristine is seen lesser and lesser after a certain probability value, and therefore we might get negatively impacted by the noise in synthetic data.

\noindent\textbf{Effect of more synthetic samples.} 
Similar to longer training, adding more synthetic samples also has a curved performance as shown in Table~\ref{tab:data}. 
Adding $1\times$ data is better than training on only real data. 
Further increasing the number of synthetic samples to $2\times$ improves the performance. 
In our setting though, increasing them more leads to a reduction in performance. 
Ideally, the goal would be to keep increasing the performance with more synthetic data but we leave that for future work as it would require looking into improving diversity of generations which will happen as the generators improve. 
Also adding more data under the current scheme ends up generating more images with the same layout leading to a drop after a certain point. 
Increasing diversity in layouts would be another way to improving performance further. 
Again, it should be noted that in all cases, our performance is sufficiently better than vanilla training.

\subsection{Qualitative Results}
We show the outputs of different parts of pipeline qualitatively on the COCO dataset. 
First, we show a few samples which are the synthetic images generated using the pre-trained grounded inpainting diffusion model. 
Figure~\ref{fig:generations} shows qualitative outputs using COCO images and annotations. 
The first column corresponds to the real image from COCO dataset followed by multiple synthetic generations each with a different seed leading to variability in instances.

We highlight a few samples which are rejected from the image level filtering step in Figure~\ref{fig:aesthetic}. 
As can be seen these are images which do not look good at an image level even though some of the instances might look fine. 
The aesthetic filtering steps removes such images. It can be seen for example in the first image in first row, the bicycle generation in the front seems incorrect. Similarly, in the first image in the second row, the bigger airplane which covers a major part of the image does not look realistic. 

Finally, in Figure~\ref{fig:det} we show some examples of instances discarded by detector filtering. 
As can be seen in most cases the detector filtering discards instances which have either been incorrectly generated or are bad quality generations. For example in the first example in the first row, the generation does not correspond to the pizza rather a specific topping. Similarly in the first example in the second row, there is no airplane generated in the ground truth region.

\section{Conclusion}
\label{sec:conclusion}

With the huge strides in image generation in terms of both quality and control, we try to tackle the problem of training detection and segmentation models with synthetic data. Our proposed pipeline Gen2Det utilizes state-of-art grounded inpainting diffusion model to generate synthetic images which we further filter at both image and instance level before using in training. We also introduce some changes in the detector training to utilize the data better and take care of shortcoming which the data might pose. Along with detailed quantitative ablations and qualitative outputs we show the efficacy of different components of our approach. Finally, we show improvement across both LVIS and COCO and show higher improvements on rare classes in the long tailed LVIS setting. Additionally, we show improvement in low-data COCO setting too. Most interestingly, we show improvement in segmentation performance without using any additional segmentation masks like existing works. We hope Gen2Det acts as a general modular framework which can be benefit from future developments in both generation as well as detector training due to easily replaceable general purpose blocks which it comprises of.

\section*{Acknowledgements}
We thank Justin Johnson for insightful discussion on this work. We also want to thank Vikas Chandra for helping with the submission.

{
    \small
    \bibliographystyle{ieeenat_fullname}
    \bibliography{main}

\begin{thebibliography}{51}
\providecommand{\natexlab}[1]{#1}
\providecommand{\url}[1]{\texttt{#1}}
\expandafter\ifx\csname urlstyle\endcsname\relax
  \providecommand{\doi}[1]{doi: #1}\else
  \providecommand{\doi}{doi: \begingroup \urlstyle{rm}\Url}\fi

\bibitem[Azizi et~al.(2023)Azizi, Kornblith, Saharia, Norouzi, and Fleet]{azizi2023synthetic}
Shekoofeh Azizi, Simon Kornblith, Chitwan Saharia, Mohammad Norouzi, and David~J Fleet.
\newblock Synthetic data from diffusion models improves imagenet classification.
\newblock \emph{arXiv preprint arXiv:2304.08466}, 2023.

\bibitem[Bansal and Grover(2023)]{bansal2023leaving}
Hritik Bansal and Aditya Grover.
\newblock Leaving reality to imagination: Robust classification via generated datasets.
\newblock \emph{arXiv preprint arXiv:2302.02503}, 2023.

\bibitem[Brooks et~al.(2023)Brooks, Holynski, and Efros]{brooks2023instructpix2pix}
Tim Brooks, Aleksander Holynski, and Alexei~A Efros.
\newblock Instructpix2pix: Learning to follow image editing instructions.
\newblock In \emph{Proceedings of the IEEE/CVF Conference on Computer Vision and Pattern Recognition}, pages 18392--18402, 2023.

\bibitem[Burgert et~al.(2022)Burgert, Ranasinghe, Li, and Ryoo]{burgert2022peekaboo}
Ryan Burgert, Kanchana Ranasinghe, Xiang Li, and Michael~S Ryoo.
\newblock Peekaboo: Text to image diffusion models are zero-shot segmentors.
\newblock \emph{arXiv preprint arXiv:2211.13224}, 2022.

\bibitem[Christophschuhmann()]{Christophschuhmann}
Christophschuhmann.
\newblock Christophschuhmann/improved-aesthetic-predictor: Clip+mlp aesthetic score predictor.

\bibitem[Clark and Jaini(2023)]{clark2023text}
Kevin Clark and Priyank Jaini.
\newblock Text-to-image diffusion models are zero-shot classifiers.
\newblock \emph{arXiv preprint arXiv:2303.15233}, 2023.

\bibitem[Deng et~al.(2009)Deng, Dong, Socher, Li, Li, and Fei-Fei]{Deng2009ImageNetAL}
Jia Deng, Wei Dong, Richard Socher, Li-Jia Li, K. Li, and Li Fei-Fei.
\newblock Imagenet: A large-scale hierarchical image database.
\newblock \emph{2009 IEEE Conference on Computer Vision and Pattern Recognition}, pages 248--255, 2009.

\bibitem[Dvornik et~al.(2018)Dvornik, Mairal, and Schmid]{Dvornik2018ModelingVC}
Nikita Dvornik, Julien Mairal, and Cordelia Schmid.
\newblock Modeling visual context is key to augmenting object detection datasets.
\newblock \emph{ArXiv}, abs/1807.07428, 2018.

\bibitem[Dwibedi et~al.(2017)Dwibedi, Misra, and Hebert]{Dwibedi2017CutPA}
Debidatta Dwibedi, Ishan Misra, and Martial Hebert.
\newblock Cut, paste and learn: Surprisingly easy synthesis for instance detection.
\newblock \emph{2017 IEEE International Conference on Computer Vision (ICCV)}, pages 1310--1319, 2017.

\bibitem[Fang et~al.(2019)Fang, Sun, Wang, Gou, Li, and Lu]{Fang2019InstaBoostBI}
Haoshu Fang, Jianhua Sun, Runzhong Wang, Minghao Gou, Yong-Lu Li, and Cewu Lu.
\newblock Instaboost: Boosting instance segmentation via probability map guided copy-pasting.
\newblock \emph{2019 IEEE/CVF International Conference on Computer Vision (ICCV)}, pages 682--691, 2019.

\bibitem[Georgakis et~al.(2017)Georgakis, Mousavian, Berg, and Kosecka]{georgakis2017synthesizing}
Georgios Georgakis, Arsalan Mousavian, Alexander~C Berg, and Jana Kosecka.
\newblock Synthesizing training data for object detection in indoor scenes.
\newblock \emph{Robotics: Science and Systems (RSS)}, 2017.

\bibitem[Ghiasi et~al.(2021)Ghiasi, Cui, Srinivas, Qian, Lin, Cubuk, Le, and Zoph]{ghiasi2021simple}
Golnaz Ghiasi, Yin Cui, Aravind Srinivas, Rui Qian, Tsung-Yi Lin, Ekin~D Cubuk, Quoc~V Le, and Barret Zoph.
\newblock Simple copy-paste is a strong data augmentation method for instance segmentation.
\newblock In \emph{Proceedings of the IEEE/CVF conference on computer vision and pattern recognition}, pages 2918--2928, 2021.

\bibitem[Gupta et~al.(2019)Gupta, Dollar, and Girshick]{gupta2019lvis}
Agrim Gupta, Piotr Dollar, and Ross Girshick.
\newblock Lvis: A dataset for large vocabulary instance segmentation.
\newblock In \emph{Proceedings of the IEEE/CVF conference on computer vision and pattern recognition}, pages 5356--5364, 2019.

\bibitem[He et~al.(2017)He, Gkioxari, Doll{\'a}r, and Girshick]{He2017MaskR}
Kaiming He, Georgia Gkioxari, Piotr Doll{\'a}r, and Ross~B. Girshick.
\newblock Mask r-cnn.
\newblock 2017.

\bibitem[He et~al.(2022)He, Sun, Yu, Xue, Zhang, Torr, Bai, and Qi]{he2022synthetic}
Ruifei He, Shuyang Sun, Xin Yu, Chuhui Xue, Wenqing Zhang, Philip Torr, Song Bai, and Xiaojuan Qi.
\newblock Is synthetic data from generative models ready for image recognition?
\newblock \emph{arXiv preprint arXiv:2210.07574}, 2022.

\bibitem[Hemmat et~al.(2023)Hemmat, Pezeshki, Bordes, Drozdzal, and Romero-Soriano]{hemmat2023feedback}
Reyhane~Askari Hemmat, Mohammad Pezeshki, Florian Bordes, Michal Drozdzal, and Adriana Romero-Soriano.
\newblock Feedback-guided data synthesis for imbalanced classification.
\newblock \emph{arXiv preprint arXiv:2310.00158}, 2023.

\bibitem[H{\"o}llein et~al.(2023)H{\"o}llein, Cao, Owens, Johnson, and Nie{\ss}ner]{hollein2023text2room}
Lukas H{\"o}llein, Ang Cao, Andrew Owens, Justin Johnson, and Matthias Nie{\ss}ner.
\newblock Text2room: Extracting textured 3d meshes from 2d text-to-image models.
\newblock \emph{arXiv preprint arXiv:2303.11989}, 2023.

\bibitem[Hu et~al.(2021)Hu, Shen, Wallis, Allen-Zhu, Li, Wang, Wang, and Chen]{hu2021lora}
Edward~J Hu, Yelong Shen, Phillip Wallis, Zeyuan Allen-Zhu, Yuanzhi Li, Shean Wang, Lu Wang, and Weizhu Chen.
\newblock Lora: Low-rank adaptation of large language models.
\newblock \emph{arXiv preprint arXiv:2106.09685}, 2021.

\bibitem[Li et~al.(2023{\natexlab{a}})Li, Prabhudesai, Duggal, Brown, and Pathak]{li2023your}
Alexander~C Li, Mihir Prabhudesai, Shivam Duggal, Ellis Brown, and Deepak Pathak.
\newblock Your diffusion model is secretly a zero-shot classifier.
\newblock \emph{arXiv preprint arXiv:2303.16203}, 2023{\natexlab{a}}.

\bibitem[Li et~al.(2023{\natexlab{b}})Li, Ling, Kar, Acuna, Kim, Kreis, Torralba, and Fidler]{li2023dreamteacher}
Daiqing Li, Huan Ling, Amlan Kar, David Acuna, Seung~Wook Kim, Karsten Kreis, Antonio Torralba, and Sanja Fidler.
\newblock Dreamteacher: Pretraining image backbones with deep generative models.
\newblock In \emph{Proceedings of the IEEE/CVF International Conference on Computer Vision}, pages 16698--16708, 2023{\natexlab{b}}.

\bibitem[Li et~al.(2023{\natexlab{c}})Li, Liu, Wu, Mu, Yang, Gao, Li, and Lee]{li2023gligen}
Yuheng Li, Haotian Liu, Qingyang Wu, Fangzhou Mu, Jianwei Yang, Jianfeng Gao, Chunyuan Li, and Yong~Jae Lee.
\newblock Gligen: Open-set grounded text-to-image generation.
\newblock In \emph{Proceedings of the IEEE/CVF Conference on Computer Vision and Pattern Recognition}, pages 22511--22521, 2023{\natexlab{c}}.

\bibitem[Lin et~al.(2023)Lin, Wang, Zeng, and Zhao]{lin2023explore}
Shaobo Lin, Kun Wang, Xingyu Zeng, and Rui Zhao.
\newblock Explore the power of synthetic data on few-shot object detection.
\newblock In \emph{Proceedings of the IEEE/CVF Conference on Computer Vision and Pattern Recognition}, pages 638--647, 2023.

\bibitem[Lin et~al.(2014)Lin, Maire, Belongie, Hays, Perona, Ramanan, Doll{\'a}r, and Zitnick]{lin2014microsoft}
Tsung-Yi Lin, Michael Maire, Serge Belongie, James Hays, Pietro Perona, Deva Ramanan, Piotr Doll{\'a}r, and C~Lawrence Zitnick.
\newblock Microsoft coco: Common objects in context.
\newblock In \emph{Computer Vision--ECCV 2014: 13th European Conference, Zurich, Switzerland, September 6-12, 2014, Proceedings, Part V 13}, pages 740--755. Springer, 2014.

\bibitem[L\"uddecke and Ecker(2022)]{lueddecke22_cvpr}
Timo L\"uddecke and Alexander Ecker.
\newblock Image segmentation using text and image prompts.
\newblock In \emph{Proceedings of the IEEE/CVF Conference on Computer Vision and Pattern Recognition (CVPR)}, pages 7086--7096, 2022.

\bibitem[Luo et~al.(2023)Luo, Dunlap, Park, Holynski, and Darrell]{luo2023dhf}
Grace Luo, Lisa Dunlap, Dong~Huk Park, Aleksander Holynski, and Trevor Darrell.
\newblock Diffusion hyperfeatures: Searching through time and space for semantic correspondence.
\newblock In \emph{Advances in Neural Information Processing Systems}, 2023.

\bibitem[Meng et~al.(2021)Meng, He, Song, Song, Wu, Zhu, and Ermon]{meng2021sdedit}
Chenlin Meng, Yutong He, Yang Song, Jiaming Song, Jiajun Wu, Jun-Yan Zhu, and Stefano Ermon.
\newblock Sdedit: Guided image synthesis and editing with stochastic differential equations.
\newblock \emph{arXiv preprint arXiv:2108.01073}, 2021.

\bibitem[Mukhopadhyay et~al.(2023)Mukhopadhyay, Gwilliam, Agarwal, Padmanabhan, Swaminathan, Hegde, Zhou, and Shrivastava]{mukhopadhyay2023diffusion}
Soumik Mukhopadhyay, Matthew Gwilliam, Vatsal Agarwal, Namitha Padmanabhan, Archana Swaminathan, Srinidhi Hegde, Tianyi Zhou, and Abhinav Shrivastava.
\newblock Diffusion models beat gans on image classification.
\newblock \emph{arXiv preprint arXiv:2307.08702}, 2023.

\bibitem[Patashnik et~al.(2023)Patashnik, Garibi, Azuri, Averbuch-Elor, and Cohen-Or]{patashnik2023localizing}
Or Patashnik, Daniel Garibi, Idan Azuri, Hadar Averbuch-Elor, and Daniel Cohen-Or.
\newblock Localizing object-level shape variations with text-to-image diffusion models.
\newblock \emph{arXiv preprint arXiv:2303.11306}, 2023.

\bibitem[Qin et~al.(2020)Qin, Zhang, Huang, Dehghan, Zaiane, and Jagersand]{Qin_2020_PR}
Xuebin Qin, Zichen Zhang, Chenyang Huang, Masood Dehghan, Osmar Zaiane, and Martin Jagersand.
\newblock U2-net: Going deeper with nested u-structure for salient object detection.
\newblock page 107404, 2020.

\bibitem[Radford et~al.(2021)Radford, Kim, Hallacy, Ramesh, Goh, Agarwal, Sastry, Askell, Mishkin, Clark, et~al.]{radford2021learning}
Alec Radford, Jong~Wook Kim, Chris Hallacy, Aditya Ramesh, Gabriel Goh, Sandhini Agarwal, Girish Sastry, Amanda Askell, Pamela Mishkin, Jack Clark, et~al.
\newblock Learning transferable visual models from natural language supervision.
\newblock In \emph{International conference on machine learning}, pages 8748--8763. PMLR, 2021.

\bibitem[Rajpura et~al.(2017)Rajpura, Hegde, and Bojinov]{Rajpura2017ObjectDU}
Param~S. Rajpura, Ravi~Sadananda Hegde, and Hristo Bojinov.
\newblock Object detection using deep cnns trained on synthetic images.
\newblock \emph{ArXiv}, abs/1706.06782, 2017.

\bibitem[Ramesh et~al.(2021)Ramesh, Pavlov, Goh, Gray, Voss, Radford, Chen, and Sutskever]{ramesh2021zero}
Aditya Ramesh, Mikhail Pavlov, Gabriel Goh, Scott Gray, Chelsea Voss, Alec Radford, Mark Chen, and Ilya Sutskever.
\newblock Zero-shot text-to-image generation.
\newblock In \emph{International Conference on Machine Learning}, pages 8821--8831. PMLR, 2021.

\bibitem[Ramesh et~al.(2022)Ramesh, Dhariwal, Nichol, Chu, and Chen]{ramesh2022hierarchical}
Aditya Ramesh, Prafulla Dhariwal, Alex Nichol, Casey Chu, and Mark Chen.
\newblock Hierarchical text-conditional image generation with clip latents.
\newblock \emph{arXiv preprint arXiv:2204.06125}, 1\penalty0 (2):\penalty0 3, 2022.

\bibitem[Ren et~al.(2015)Ren, He, Girshick, and Sun]{Ren2015FasterRT}
Shaoqing Ren, Kaiming He, Ross~B. Girshick, and Jian Sun.
\newblock Faster r-cnn: Towards real-time object detection with region proposal networks.
\newblock \emph{IEEE Transactions on Pattern Analysis and Machine Intelligence}, 39:\penalty0 1137--1149, 2015.

\bibitem[Rombach et~al.(2022)Rombach, Blattmann, Lorenz, Esser, and Ommer]{rombach2022high}
Robin Rombach, Andreas Blattmann, Dominik Lorenz, Patrick Esser, and Bj{\"o}rn Ommer.
\newblock High-resolution image synthesis with latent diffusion models.
\newblock In \emph{Proceedings of the IEEE/CVF conference on computer vision and pattern recognition}, pages 10684--10695, 2022.

\bibitem[Ruiz et~al.(2023)Ruiz, Li, Jampani, Pritch, Rubinstein, and Aberman]{ruiz2023dreambooth}
Nataniel Ruiz, Yuanzhen Li, Varun Jampani, Yael Pritch, Michael Rubinstein, and Kfir Aberman.
\newblock Dreambooth: Fine tuning text-to-image diffusion models for subject-driven generation.
\newblock In \emph{Proceedings of the IEEE/CVF Conference on Computer Vision and Pattern Recognition}, pages 22500--22510, 2023.

\bibitem[Saharia et~al.(2022{\natexlab{a}})Saharia, Chan, Chang, Lee, Ho, Salimans, Fleet, and Norouzi]{saharia2022palette}
Chitwan Saharia, William Chan, Huiwen Chang, Chris Lee, Jonathan Ho, Tim Salimans, David Fleet, and Mohammad Norouzi.
\newblock Palette: Image-to-image diffusion models.
\newblock In \emph{ACM SIGGRAPH 2022 Conference Proceedings}, pages 1--10, 2022{\natexlab{a}}.

\bibitem[Saharia et~al.(2022{\natexlab{b}})Saharia, Chan, Saxena, Li, Whang, Denton, Ghasemipour, Gontijo~Lopes, Karagol~Ayan, Salimans, et~al.]{saharia2022photorealistic}
Chitwan Saharia, William Chan, Saurabh Saxena, Lala Li, Jay Whang, Emily~L Denton, Kamyar Ghasemipour, Raphael Gontijo~Lopes, Burcu Karagol~Ayan, Tim Salimans, et~al.
\newblock Photorealistic text-to-image diffusion models with deep language understanding.
\newblock \emph{Advances in Neural Information Processing Systems}, 35:\penalty0 36479--36494, 2022{\natexlab{b}}.

\bibitem[Sariyildiz et~al.(2023)Sariyildiz, Alahari, Larlus, and Kalantidis]{sariyildiz2023fake}
Mert~Bulent Sariyildiz, Karteek Alahari, Diane Larlus, and Yannis Kalantidis.
\newblock Fake it till you make it: Learning transferable representations from synthetic imagenet clones.
\newblock In \emph{Proceedings of the IEEE/CVF Conference on Computer Vision and Pattern Recognition (CVPR)}, 2023.

\bibitem[Su et~al.(2022)Su, Deng, Sun, Lin, and Wu]{2203.04708}
Yukun Su, Jingliang Deng, Ruizhou Sun, Guosheng Lin, and Qingyao Wu.
\newblock A unified transformer framework for group-based segmentation: Co-segmentation, co-saliency detection and video salient object detection, 2022.

\bibitem[Tang et~al.(2023)Tang, Jia, Wang, Phoo, and Hariharan]{tang2023emergent}
Luming Tang, Menglin Jia, Qianqian Wang, Cheng~Perng Phoo, and Bharath Hariharan.
\newblock Emergent correspondence from image diffusion.
\newblock \emph{arXiv preprint arXiv:2306.03881}, 2023.

\bibitem[Tian et~al.(2023)Tian, Fan, Isola, Chang, and Krishnan]{tian2023stablerep}
Yonglong Tian, Lijie Fan, Phillip Isola, Huiwen Chang, and Dilip Krishnan.
\newblock Stablerep: Synthetic images from text-to-image models make strong visual representation learners.
\newblock \emph{arXiv preprint arXiv:2306.00984}, 2023.

\bibitem[Tremblay et~al.(2018)Tremblay, Prakash, Acuna, Brophy, Jampani, Anil, To, Cameracci, Boochoon, and Birchfield]{Tremblay2018TrainingDN}
Jonathan Tremblay, Aayush Prakash, David Acuna, Mark Brophy, V. Jampani, Cem Anil, Thang To, Eric Cameracci, Shaad Boochoon, and Stan Birchfield.
\newblock Training deep networks with synthetic data: Bridging the reality gap by domain randomization.
\newblock \emph{2018 IEEE/CVF Conference on Computer Vision and Pattern Recognition Workshops (CVPRW)}, pages 1082--10828, 2018.

\bibitem[Voynov et~al.(2023)Voynov, Aberman, and Cohen-Or]{voynov2023sketch}
Andrey Voynov, Kfir Aberman, and Daniel Cohen-Or.
\newblock Sketch-guided text-to-image diffusion models.
\newblock In \emph{ACM SIGGRAPH 2023 Conference Proceedings}, pages 1--11, 2023.

\bibitem[Wu et~al.(2019)Wu, Kirillov, Massa, Lo, and Girshick]{wu2019detectron2}
Yuxin Wu, Alexander Kirillov, Francisco Massa, Wan-Yen Lo, and Ross Girshick.
\newblock Detectron2.
\newblock \url{https://github.com/facebookresearch/detectron2}, 2019.

\bibitem[Xiang et~al.(2023)Xiang, Yang, Huang, and Wang]{xiang2023denoising}
Weilai Xiang, Hongyu Yang, Di Huang, and Yunhong Wang.
\newblock Denoising diffusion autoencoders are unified self-supervised learners.
\newblock \emph{arXiv preprint arXiv:2303.09769}, 2023.

\bibitem[Yu et~al.(2022)Yu, Xu, Koh, Luong, Baid, Wang, Vasudevan, Ku, Yang, Ayan, et~al.]{yu2022scaling}
Jiahui Yu, Yuanzhong Xu, Jing~Yu Koh, Thang Luong, Gunjan Baid, Zirui Wang, Vijay Vasudevan, Alexander Ku, Yinfei Yang, Burcu~Karagol Ayan, et~al.
\newblock Scaling autoregressive models for content-rich text-to-image generation.
\newblock \emph{arXiv preprint arXiv:2206.10789}, 2\penalty0 (3):\penalty0 5, 2022.

\bibitem[Yun and Lin(2022)]{yun2022selfreformer}
Yi~Ke Yun and Weisi Lin.
\newblock Selfreformer: Self-refined network with transformer for salient object detection.
\newblock \emph{arXiv preprint arXiv:2205.11283}, 2022.

\bibitem[Zhang et~al.(2023)Zhang, Rao, and Agrawala]{zhang2023adding}
Lvmin Zhang, Anyi Rao, and Maneesh Agrawala.
\newblock Adding conditional control to text-to-image diffusion models.
\newblock In \emph{Proceedings of the IEEE/CVF International Conference on Computer Vision}, pages 3836--3847, 2023.

\bibitem[Zhao et~al.(2023)Zhao, Sheng, Bao, Chen, Chen, Wen, Yuan, Liu, Zhou, Chu, et~al.]{zhao2023x}
Hanqing Zhao, Dianmo Sheng, Jianmin Bao, Dongdong Chen, Dong Chen, Fang Wen, Lu Yuan, Ce Liu, Wenbo Zhou, Qi Chu, et~al.
\newblock X-paste: Revisiting scalable copy-paste for instance segmentation using clip and stablediffusion.
\newblock 2023.

\bibitem[Zhou et~al.(2021)Zhou, Koltun, and Kr{\"a}henb{\"u}hl]{zhou2021probabilistic}
Xingyi Zhou, Vladlen Koltun, and Philipp Kr{\"a}henb{\"u}hl.
\newblock Probabilistic two-stage detection.
\newblock \emph{arXiv preprint arXiv:2103.07461}, 2021.

\end{thebibliography}
}

\clearpage
\maketitlesupplementary

\begin{table}
\setlength{\cmidrulewidth}{0.01em}
\renewcommand{\tabcolsep}{10pt}
\renewcommand{\arraystretch}{1.1}
\centering
\resizebox{0.7\linewidth}{!}{
\begin{tabular}{@{}lcc@{}}
\toprule
 Method &  $\text{AP}_b$ & $\text{AP}_m$ \\
 \midrule

  Faster R-CNN & 	40.20 & -\\
 Faster R-CNN (Ours) &\textbf{40.56}	&-\\
 
 \hdashline
  Mask R-CNN & 	41.08	& 37.14\\
 Mask R-CNN (Ours) &\textbf{41.53}	&\textbf{37.46}\\
 
 \hdashline
  CenterNet2 & 	46.00	& 39.8\\
 CenterNet2 (Ours) &\textbf{47.18}	& \textbf{40.44}\\
\bottomrule
\end{tabular}
}
\caption{Comparisons across architectures on COCO dataset. We report the Box AP ($\text{AP}_b$) and Mask AP ($\text{AP}_m$).}\label{tab:coco_arch}
\vspace{-0.2in}
\end{table}

\section{Experimental Details}
In addition to the details mentioned in the main paper we would like to highlight that we were not able to reproduce the exact results of XPaste from the code and configs provided and it is an \href{https://github.com/yoctta/XPaste/issues/2}{open issue} in their Github repository. Additionally, for learning rate for Faster and Mask R-CNN experiments we use the default LR of 0.08 which is scaled from 0.02 as we use $4\times$ the batch size as the default configs. For LVIS, this LR was not stable so we use an LR of 0.04 accross all Faster and Mask R-CNN experiments. For LD-COCO we do not perform filtering or background ignore due to already low number of images and instances.

\section{Quantitative Results on COCO}
In Table~\ref{tab:coco_arch} we show improvements in both Box ($\text{AP}_b$) and Mask ($\text{AP}_m$) AP across Faster R-CNN, Mask R-CNN and Centernet2 architectures on the COCO dataset.

\section{Pasting Object Centric Instances}
We use state-of-art text to image diffusion model to generate object centric images for each instance annotation and paste them on corresponding LVIS images in a layout satisfying manner using the box coordinates. We use the prompt $\text{a photo of }<c>$ where $c$ is the class name for that instance. We then train the detector with vanilla training and sampling synthetic and real data with equal probabilities. This gives us a Box AP of $22.50$ and Mask AP of $21.45$ which is lower than the performance obtained through Gen2Det where we get $24.42$ Box AP and $23.67$ Mask AP.

\section{Ablations}
\subsection{Ablation on $\tau_s$}
We vary $\tau_s$ which is the score threshold we use for detector filtering. For this experiment we keep $\tau_{iou}$ fixed at 0.3. We show the results for this in Table~\ref{tab:det_filt_taus}. It can be seen that initially increasing the score threshold helps but after 0.2 increasing it further leads to a reduction in performance. This might be because we filter out too many annotations.

\subsection{Ablation on $\tau_{iou}$}
We vary $\tau_{iou}$ which is the score threshold we use for detector filtering. For this experiment we keep $\tau_{s}$ fixed at 0.2. We show the results for this in Table~\ref{tab:det_filt_tauiou}. It can be seen that initially increasing the IoU threshold helps but after 0.3 increasing it further leads to a reduction in performance. This might be because we filter out too many annotations.

\subsection{Ablation on $\tau_{i}$}
We vary $\tau_{i}$ which is the score threshold we use for background ignore. We show the results for this in Table~\ref{tab:ignore_thresh}. It can be seen that initially setting the score threshold to 0 which ends up ignoring all background instances from the synthetic data works the best. After that increasing it reduces it with sporadic increases at certain thresholds.

\section{Qualitative Results}
\subsection{Qualitative Inpainted Generation}
\begin{figure*}
\centering
\includegraphics[width=\linewidth]{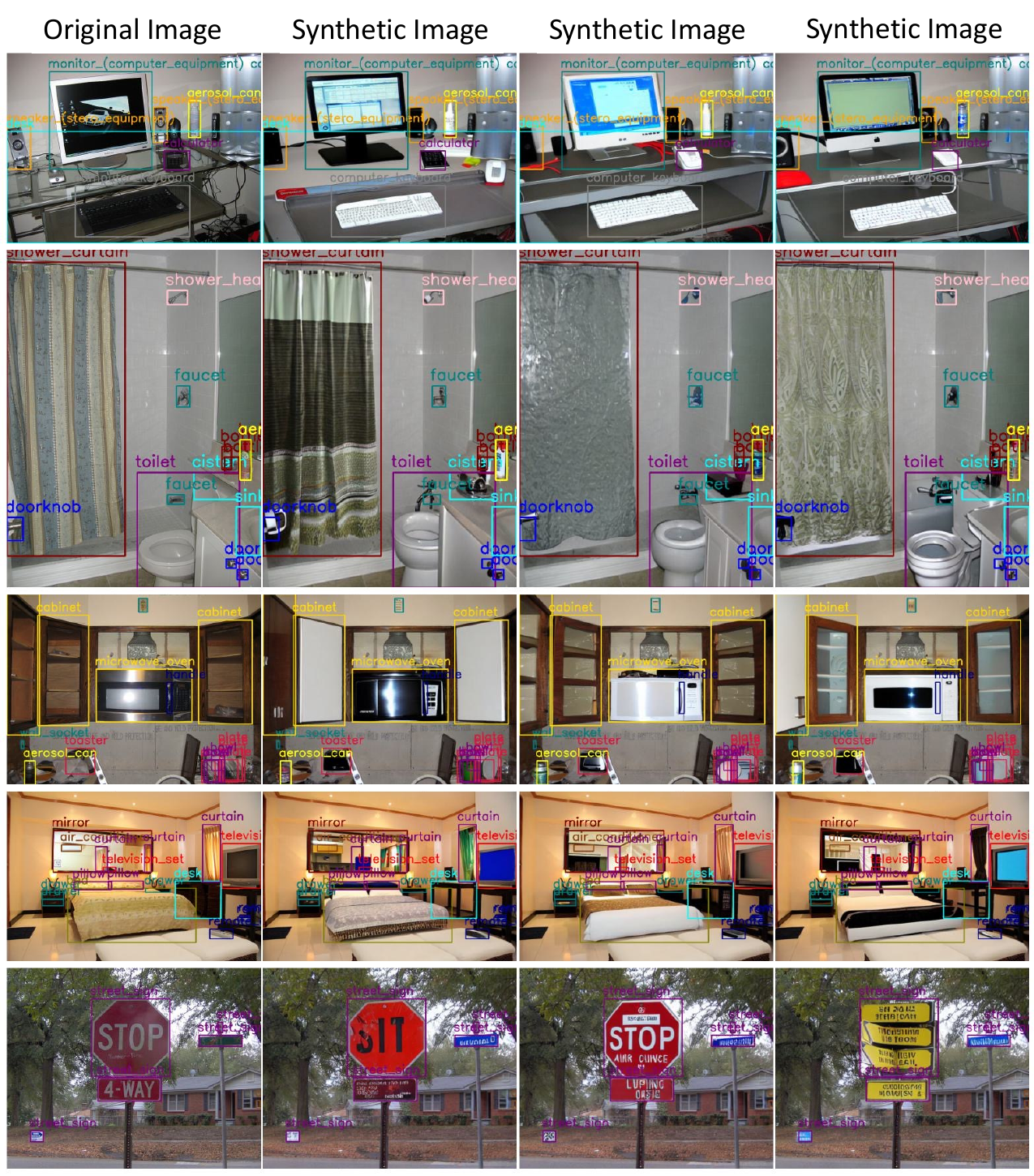}
  \caption{
  Examples of generations using the inpainting diffusion model on the LVIS dataset. 
  The first column corresponds to the original LVIS images and the rest of the columns are generations with different seeds.
  These generated images are then fed to our filtering pipeline for further processing before use in training. 
  }
\label{fig:lvis_generations_supp_1}
\vspace{-0.2in}
\end{figure*}

\begin{figure*}
\centering
\includegraphics[width=\linewidth]{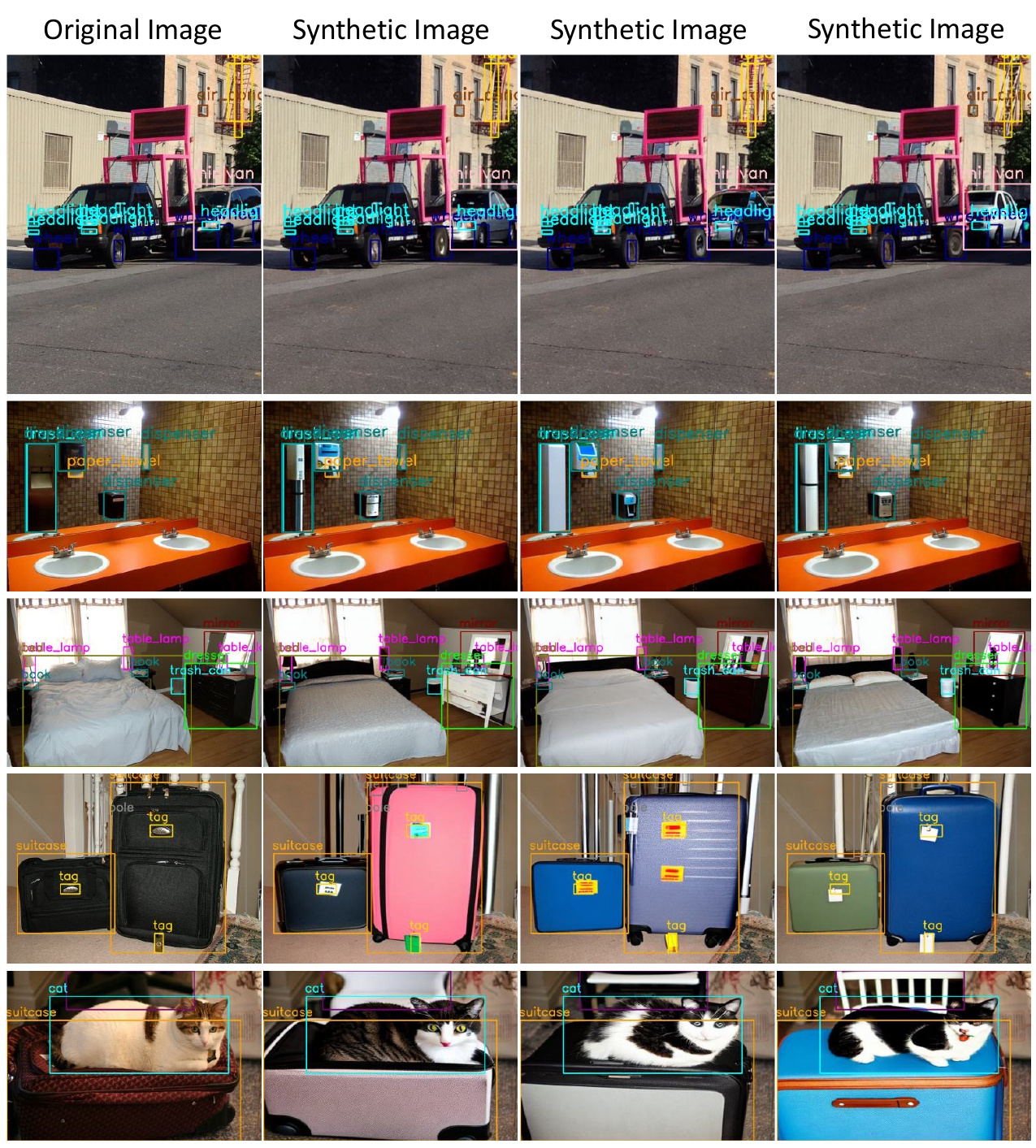}
  \caption{
  Examples of generations using the inpainting diffusion model on the LVIS dataset. 
  The first column corresponds to the original LVIS images and the rest of the columns are generations with different seeds.
  These generated images are then fed to our filtering pipeline for further processing before use in training. 
  }
\label{fig:lvis_generations_supp_2}
\vspace{-0.2in}
\end{figure*}

We show qualitative generation on the LVIS dataset in Figures~\ref{fig:lvis_generations_supp_1} and ~\ref{fig:lvis_generations_supp_2}. These generations utilize the LVIS annotations and images along with state-of-the-art grounded inpainting model to generate the instances conditioned on the class labels, box labels, and input images.

\begin{figure*}
\centering
\includegraphics[width=\linewidth]{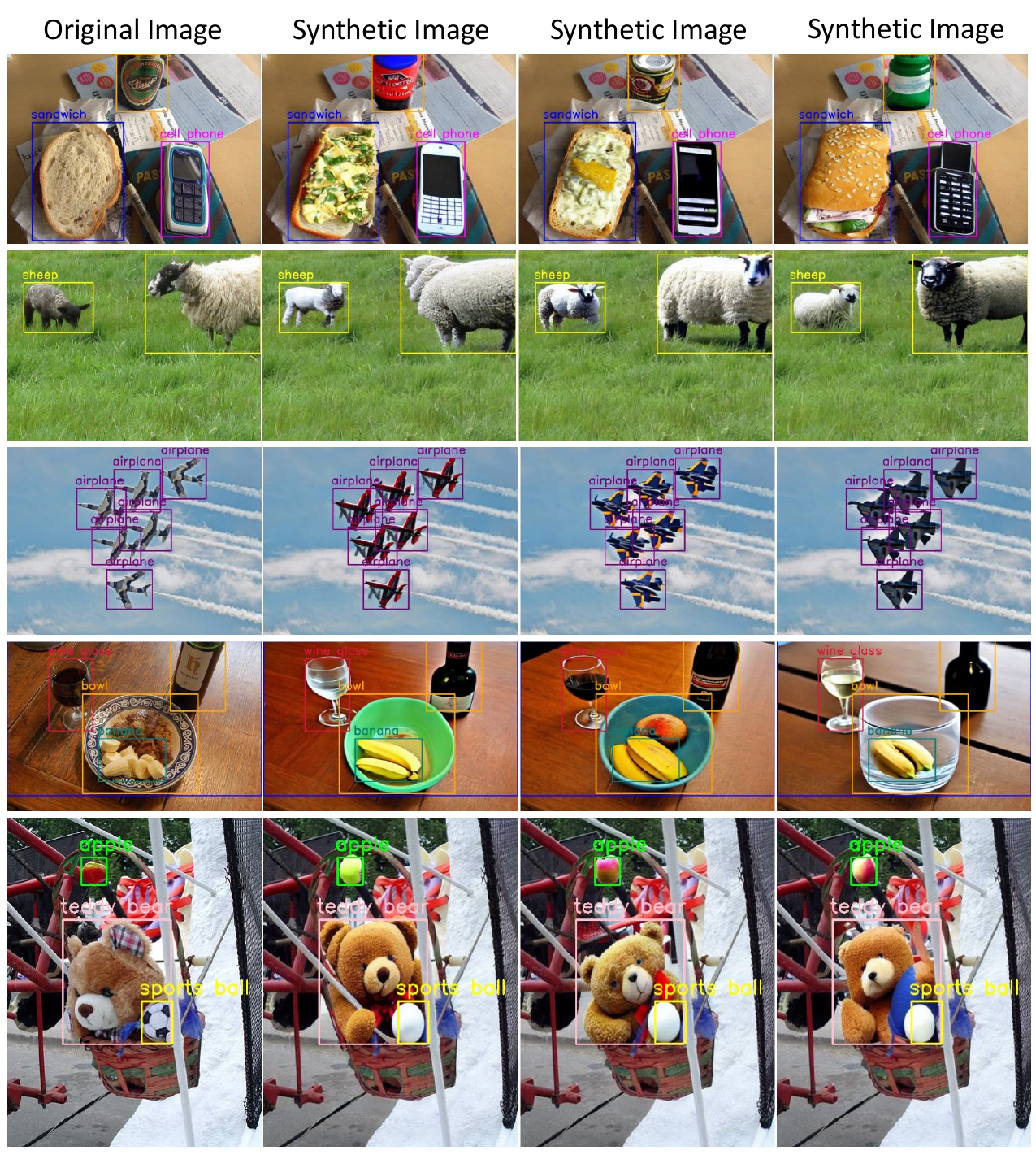}
  \caption{
  Examples of generations using the inpainting diffusion model on the COCO dataset. 
  The first column corresponds to the original COCO images and the rest of the columns are generations with different seeds.
  These generated images are then fed to our filtering pipeline for further processing before use in training. 
  }
\label{fig:coco_generations_supp_1}
\vspace{-0.2in}
\end{figure*}

\begin{figure*}
\centering
\includegraphics[width=\linewidth]{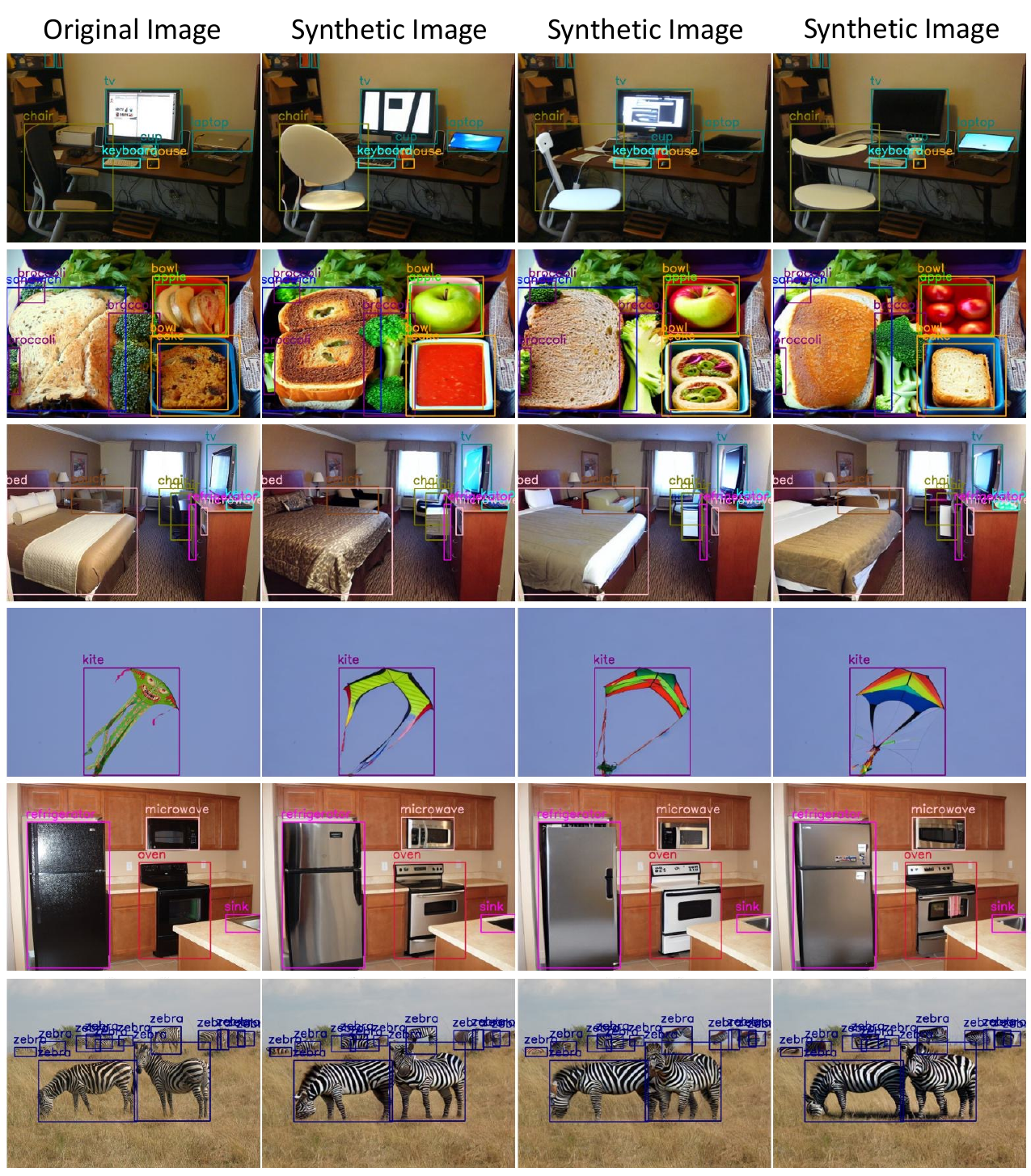}
  \caption{
  Examples of generations using the inpainting diffusion model on the COCO dataset. 
  The first column corresponds to the original COCO images and the rest of the columns are generations with different seeds.
  These generated images are then fed to our filtering pipeline for further processing before use in training. 
  }
\label{fig:coco_generations_supp_2}
\end{figure*}

We also show qualitative generation on the COCO dataset in Figures~\ref{fig:coco_generations_supp_1} and ~\ref{fig:coco_generations_supp_2} utilizing COCO annotations and images to generate them.

\subsection{Qualitative Results for Image Filtering}
\begin{figure*}
\centering
\includegraphics[width=\linewidth]{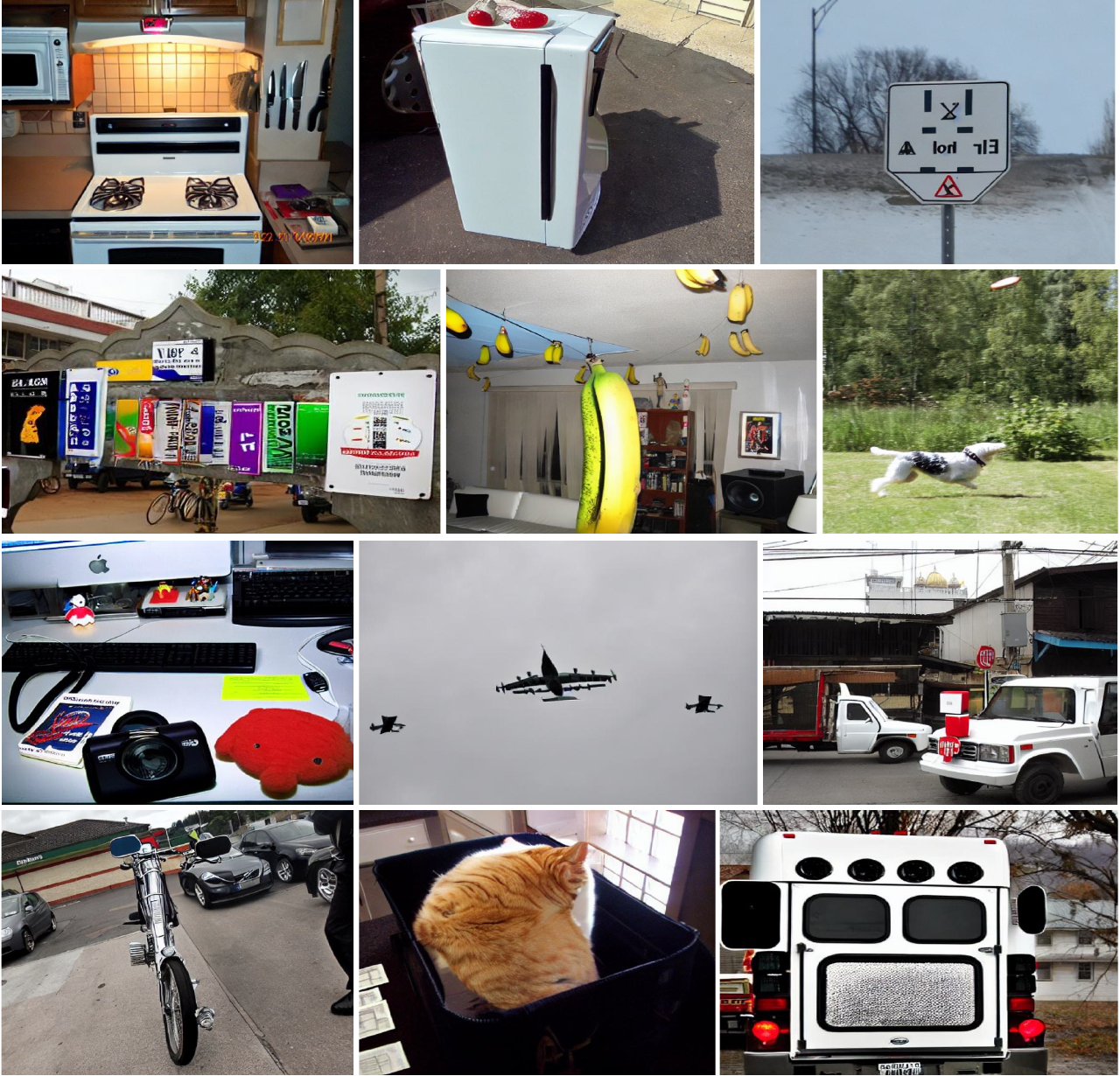}
  \caption{
  Examples of samples discarded during the image level filtering on LVIS. 
  As can be seen, image level filtering is able to remove images with artifacts present at a global level.
  }
\label{fig:image_filt_sup_lvis}
\end{figure*}

In Figure~\ref{fig:image_filt_sup_lvis} we show results of the image level filtering by visualizing images discarded during this step on the LVIS dataset. It can be seen there are high level issues in the generations. In the third image in row 1 the sign seems to have some weird patterns. In row 3 the second image has low quality aircraft generations. Similarly in the second image in the last row the cat generation is not great.

\begin{figure*}
\centering
\includegraphics[width=\linewidth]{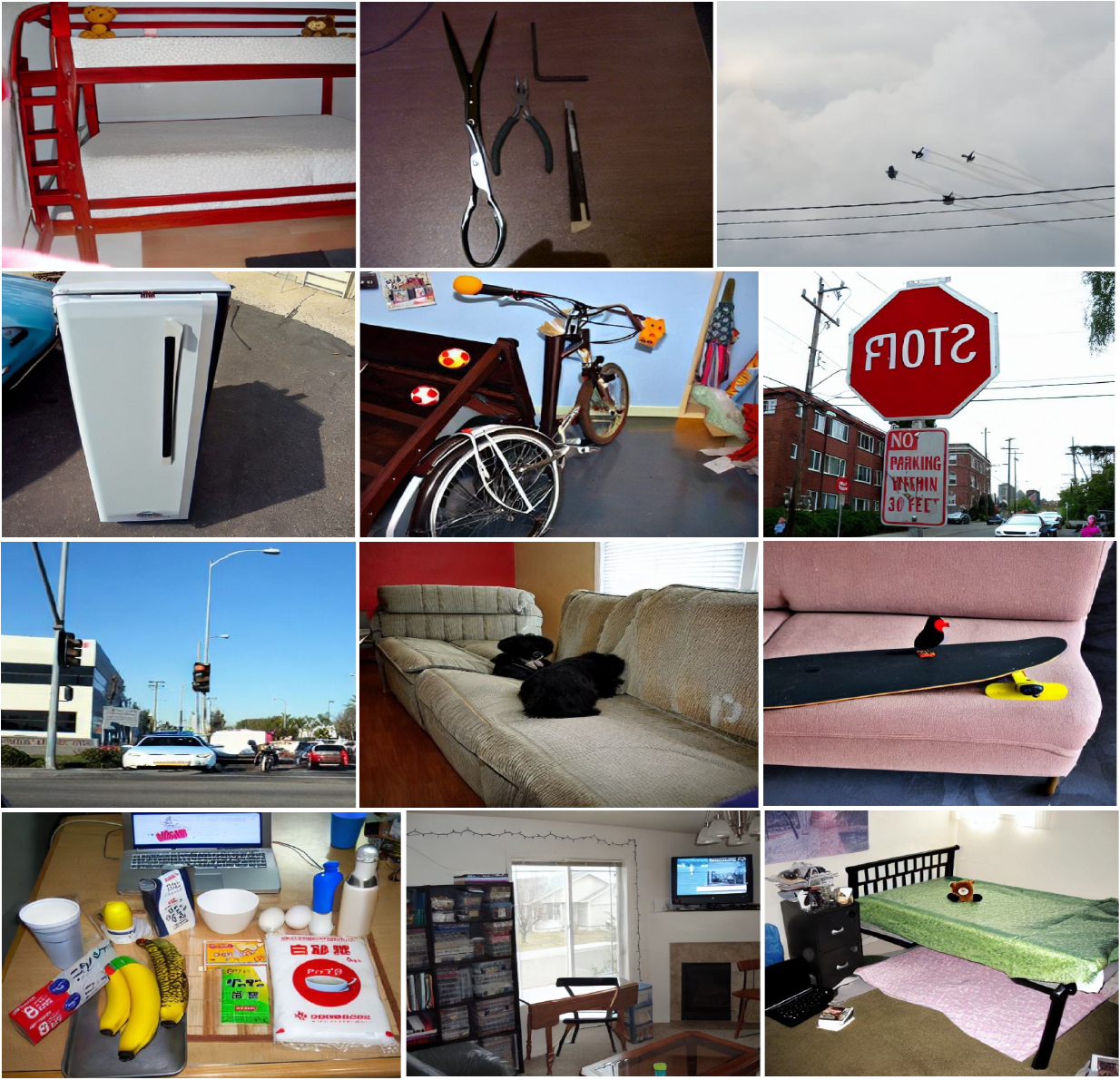}
  \caption{
  Examples of samples discarded during the image level filtering on COCO. 
  As can be seen, image level filtering is able to remove images with artifacts present at a global level.
  }
\label{fig:image_filt_sup_coco}
\end{figure*}

We also show more image filtering results on the COCO dataset and visualize the discarded images in Figure~\ref{fig:image_filt_sup_coco}.

\begin{table*}
 \setlength{\cmidrulewidth}{0.01em}
\renewcommand{\tabcolsep}{12pt}
\renewcommand{\arraystretch}{1.0}
\centering
\resizebox{0.9\linewidth}{!}{
\begin{tabular}{@{}ccccccccc@{}}
\toprule
 \multirow{2}{*}{$\tau_s$} &  \multicolumn{4}{c}{Box}&  \multicolumn{4}{c}{Mask}\\
 \cmidrule[\cmidrulewidth](l){2-5}
 \cmidrule[\cmidrulewidth](l){6-9}

 &  $\text{AP}$ & $\text{AP}_r$ & $\text{AP}_c$ & $\text{AP}_f$ &  $\text{AP}$ & $\text{AP}_r$ & $\text{AP}_c$ & $\text{AP}_f$ \\
 \midrule

0.1	&23.85&	14.04&	21.89	&30.36	&23.16&	14.21	&21.92	&28.49\\
0.2 &  \textbf{24.42} & \textbf{15.43} &\textbf{ 22.63} & \textbf{30.38} & \textbf{23.67} &\textbf{ 15.33} & \textbf{22.62} & \textbf{28.51}\\
0.3 &	23.59&	13.21&	21.74	&30.21&	22.89	&13.50	&21.66	&28.39\\
0.4	&23.75&	13.48&	21.97&	30.25	&23.03&	13.81	&21.86	&28.38\\
0.5	&23.62&	13.69&	21.56	&30.28	&23.05	&14.24&	21.75	&28.36\\
 
\bottomrule
\end{tabular}
}
\caption{
Comparisons across different $\tau_s$ keeping $\tau_{iou}$ fixed at 0.3 on the LVIS dataset. We report the overall AP and also AP for rare ($\text{AP}_r$), common ($\text{AP}_c$) and frequent ($\text{AP}_f$) classes for both box and mask evaluations.}
\label{tab:det_filt_taus}
\vspace{-0.1in}
\end{table*}

\begin{table*}
 \setlength{\cmidrulewidth}{0.01em}
\renewcommand{\tabcolsep}{12pt}
\renewcommand{\arraystretch}{1.0}
\centering
\resizebox{0.9\linewidth}{!}{
\begin{tabular}{@{}ccccccccc@{}}
\toprule
 \multirow{2}{*}{$\tau_{iou}$} &  \multicolumn{4}{c}{Box}&  \multicolumn{4}{c}{Mask}\\
 \cmidrule[\cmidrulewidth](l){2-5}
 \cmidrule[\cmidrulewidth](l){6-9}

 &  $\text{AP}$ & $\text{AP}_r$ & $\text{AP}_c$ & $\text{AP}_f$ &  $\text{AP}$ & $\text{AP}_r$ & $\text{AP}_c$ & $\text{AP}_f$ \\
 \midrule

0.1	&	23.49	&12.90	&21.70&	30.15&	22.82	&12.90&	21.77&	28.35\\
0.2	&23.56&	12.32&	22.05	&30.18&	22.88&	12.58&	22.08	&28.31\\
0.3 &  \textbf{24.42} & \textbf{15.43} &\textbf{ 22.63} & \textbf{30.38} & \textbf{23.67} &\textbf{ 15.33} & \textbf{22.62} & \textbf{28.51}\\

0.4 &23.73&	13.09&	22.02&	30.32&	23.12&	13.40&	22.13	&28.49\\
0.5	&23.77	&13.22&	21.89&	30.51	&22.91&	13.33	&21.58&	28.60\\
 
\bottomrule
\end{tabular}
}
\caption{
Comparisons across different $\tau_{iou}$ keeping $\tau_s$ fixed at 0.2 on the LVIS dataset. We report the overall AP and also AP for rare ($\text{AP}_r$), common ($\text{AP}_c$) and frequent ($\text{AP}_f$) classes for both box and mask evaluations.}
\label{tab:det_filt_tauiou}
\vspace{-0.2in}
\end{table*}

\subsection{Qualitative Results for Detector Filtering}

\begin{figure*}
\centering
\includegraphics[width=\linewidth]{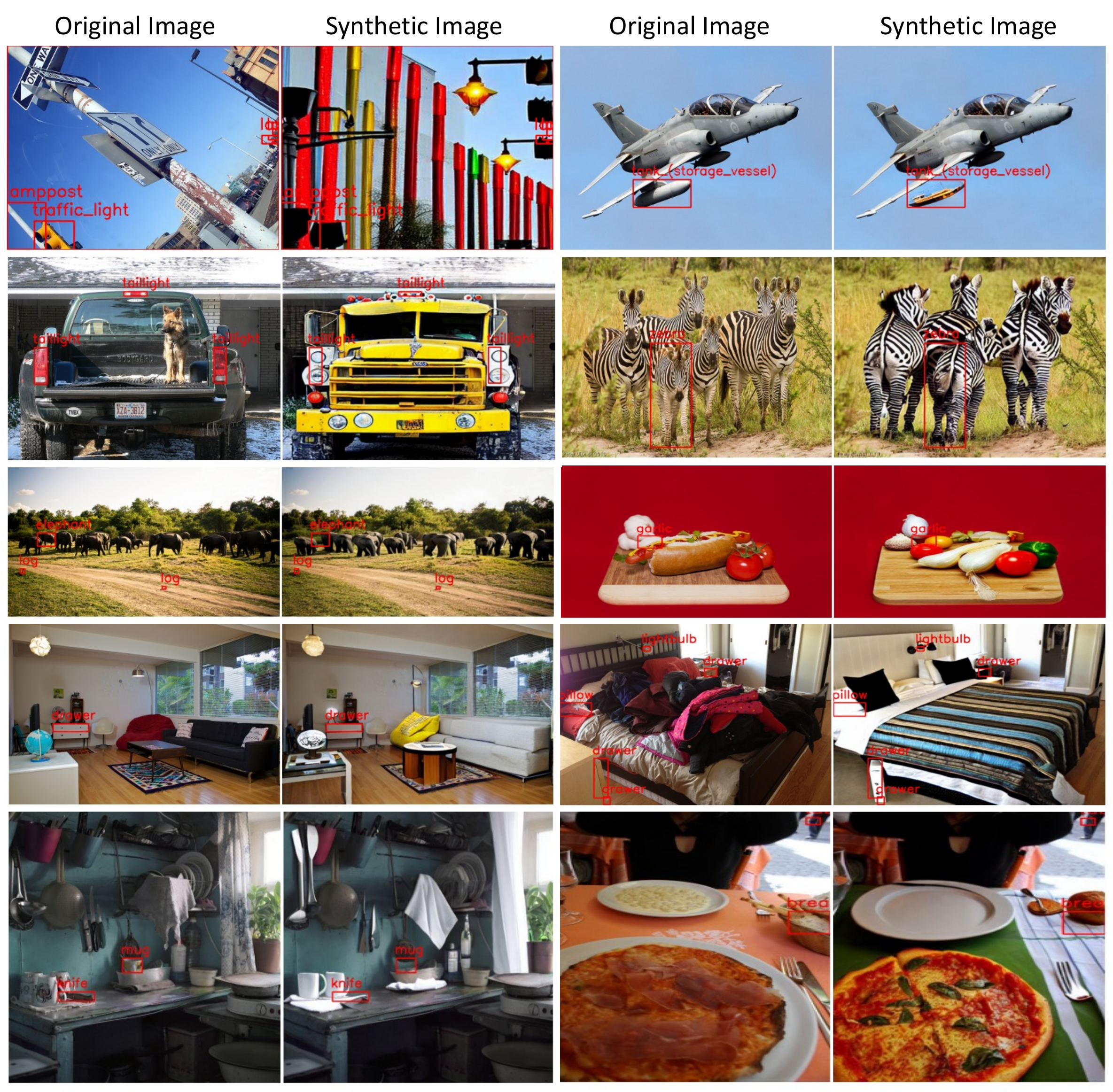}
  \caption{
  Examples of ground-truth instance annotations discarded by the detector filtering highlighted in red for LVIS dataset.
  }
\label{fig:det_filt_sup_lvis}
\end{figure*}

\begin{figure*}
\centering
\includegraphics[width=\linewidth]{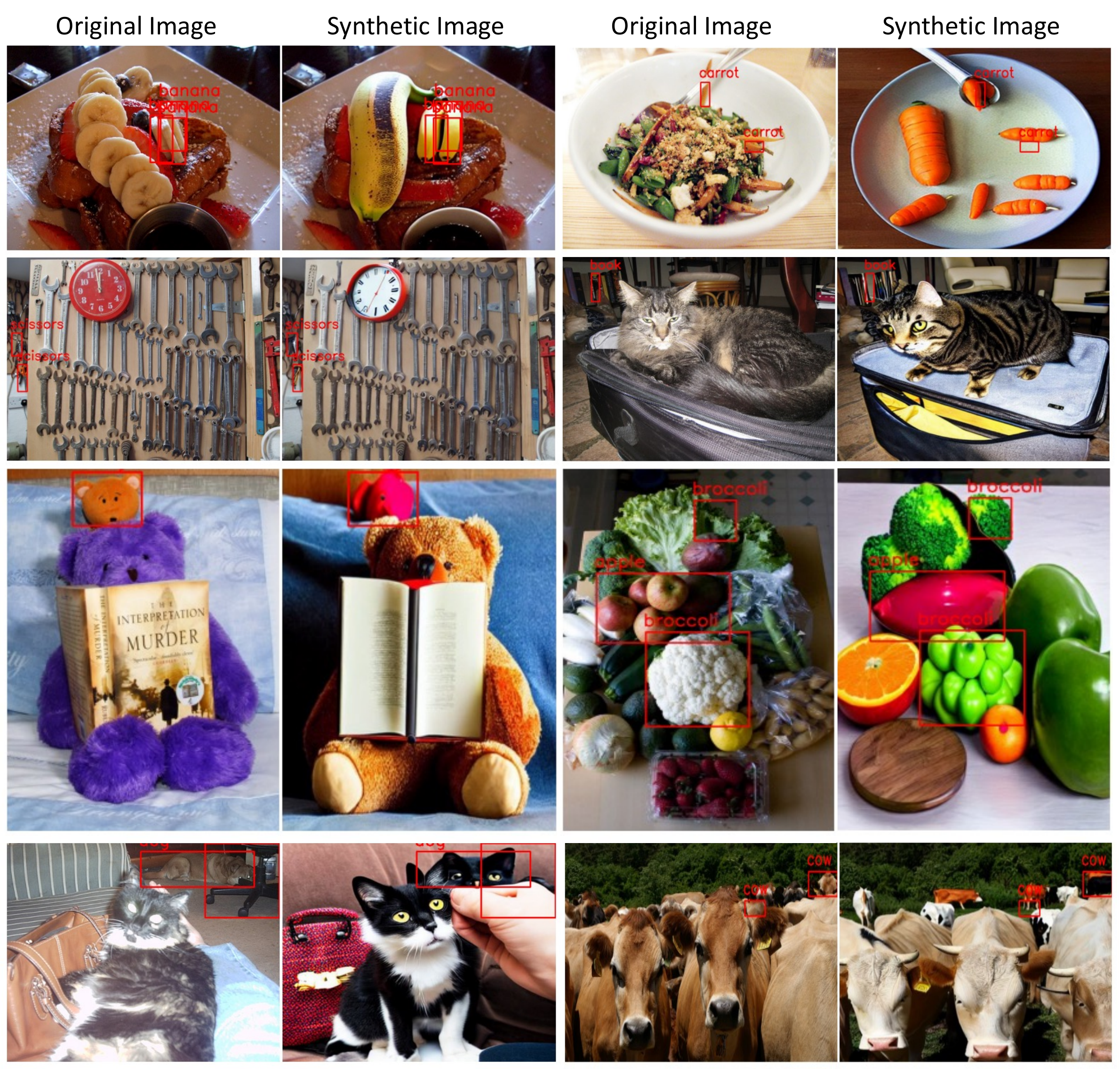}
  \caption{
  Examples of ground-truth instance annotations discarded by the detector filtering highlighted in red for COCO dataset.
  }
\label{fig:det_filt_sup_coco}
\end{figure*}

We present some examples for detector filtering on LVIS in Figure~\ref{fig:det_filt_sup_lvis}. We show the original and generated instances for each of the images and show the ground truth instance annotation(s) being discarded for each image in red. 
We also show examples of detector filtering on COCO in Figure~\ref{fig:det_filt_sup_coco}.

\begin{table*}
 \setlength{\cmidrulewidth}{0.01em}
\renewcommand{\tabcolsep}{12pt}
\renewcommand{\arraystretch}{1.0}
\centering
\resizebox{0.9\linewidth}{!}{
\begin{tabular}{@{}ccccccccc@{}}
\toprule
 \multirow{2}{*}{$\tau_{i}$} &  \multicolumn{4}{c}{Box}&  \multicolumn{4}{c}{Mask}\\
 \cmidrule[\cmidrulewidth](l){2-5}
 \cmidrule[\cmidrulewidth](l){6-9}

 &  $\text{AP}$ & $\text{AP}_r$ & $\text{AP}_c$ & $\text{AP}_f$ &  $\text{AP}$ & $\text{AP}_r$ & $\text{AP}_c$ & $\text{AP}_f$ \\
 \midrule

0.0 &  \textbf{24.42} & \textbf{15.43} &\textbf{ 22.63} & \textbf{30.38} & \textbf{23.67} &\textbf{ 15.33} & \textbf{22.62} & \textbf{28.51}\\

0.1	&23.89	&14.06&	21.88&	30.46	&23.21&	14.27	&21.92	&28.57\\
0.2	&23.92	&13.82&	22.08	&30.41	&23.34&	14.63	&22.14	&28.50\\
0.3	& 23.70	&13.27	&21.86	&30.34&	23.02&	13.32&	21.95	&28.49\\
0.4	&24.06	&14.83	&22.01&	30.39&	23.38	&15.15&	22.07	&28.45\\
0.5	&23.69	&12.31	&22.09	&30.46&	22.99&	12.61&	22.06&	28.60\\
 
\bottomrule
\end{tabular}
}
\caption{
Comparisons across different $\tau_i$ on the LVIS dataset. We report the overall AP and also AP for rare ($\text{AP}_r$), common ($\text{AP}_c$) and frequent ($\text{AP}_f$) classes for both box and mask evaluations.}
\label{tab:ignore_thresh}
\vspace{-0.2in}
\end{table*}

\end{document}